\documentclass[10pt,twocolumn,letterpaper]{article}
\pdfoutput=1
\usepackage{iccv}
\usepackage{times}
\usepackage{epsfig}
\usepackage{graphicx}
\usepackage{amsmath}
\usepackage{amssymb}
\usepackage{booktabs}
\usepackage{fixltx2e}
\usepackage{subcaption}
\usepackage{xspace}
\usepackage{soul}
\usepackage{array,multirow}
\usepackage{makecell}

\usepackage[usenames,dvipsnames, table]{xcolor}

\usepackage[pagebackref=true,breaklinks=true,letterpaper=true,colorlinks,bookmarks=false]{hyperref}

\iccvfinalcopy % *** Uncomment this line for the final submission

\newcommand{\figref}[1]{Fig.~\ref{#1}}
\newcommand{\tabref}[1]{Tab.~\ref{#1}}
\newcommand{\secref}[1]{Sec.~\ref{#1}}

\def\am{\mbox{AM}\xspace}
\def\amtv{\mbox{AM\textsubscript{TV}}\xspace}
\def\amda{\mbox{AMDA}\xspace}
\def\amdatv{\mbox{AMDA\textsubscript{TV}}\xspace}

\def\amh{\mbox{AM-\it{ft\textsubscript{Gauss}}}\xspace}

\def\amtvh{\mbox{AM\textsubscript{TV}-\it{ft\textsubscript{Gauss}}}\xspace}

\def\aml{\mbox{AM-\it{ft\textsubscript{Cont}}}\xspace}

\def\amdatvh{\mbox{AMDA\textsubscript{TV}-\it{ft\textsubscript{Gauss}}}\xspace}
\def\amdal{\mbox{AMDA-\it{ft\textsubscript{Cont}}}\xspace}

\def\rohl{\mbox{RoHL}\xspace}

\def\iid{{i.i.d}\xspace}
\def\ood{{OOD}\xspace}

\begin{document}

%%%%%%%%% TITLE
\title{Improving robustness against common corruptions with frequency biased models}

\author{Tonmoy Saikia\\
 University of Freiburg\\
{\tt\small saikiat@cs.uni-freiburg.de}
\and
Cordelia Schmid\\
Inria\\
{\tt\small cordelia.schmid@inria.fr}
\and
Thomas Brox\\
University of Freiburg\\
{\tt\small brox@cs.uni-freiburg.de}
\and
}

\maketitle
%%%%%%%%% ABSTRACT
\begin{abstract}
CNNs perform remarkably well when the training and test distributions are \iid, but unseen image corruptions can cause a surprisingly large drop in performance. In various real scenarios, unexpected distortions, such as random noise, compression artefacts or weather distortions are common phenomena. Improving performance on corrupted images must not result in degraded \iid performance -- a challenge faced by many state-of-the-art robust approaches. Image corruption types have different characteristics in the frequency spectrum and would benefit from a targeted type of data augmentation, which, however, is often unknown during training. In this paper, we introduce a mixture of two expert models specializing in high and low-frequency robustness, respectively. Moreover, we propose a new regularization scheme that minimizes the total variation (TV) of convolution feature-maps to increase high-frequency robustness. The approach improves on corrupted images without degrading in-distribution performance.
We demonstrate this on ImageNet-C and also for real-world corruptions on an automotive dataset, both for object classification and object detection.
\end{abstract}
\section{Introduction}

\begin{figure}[h]
\begin{center}
\includegraphics[width=\columnwidth]{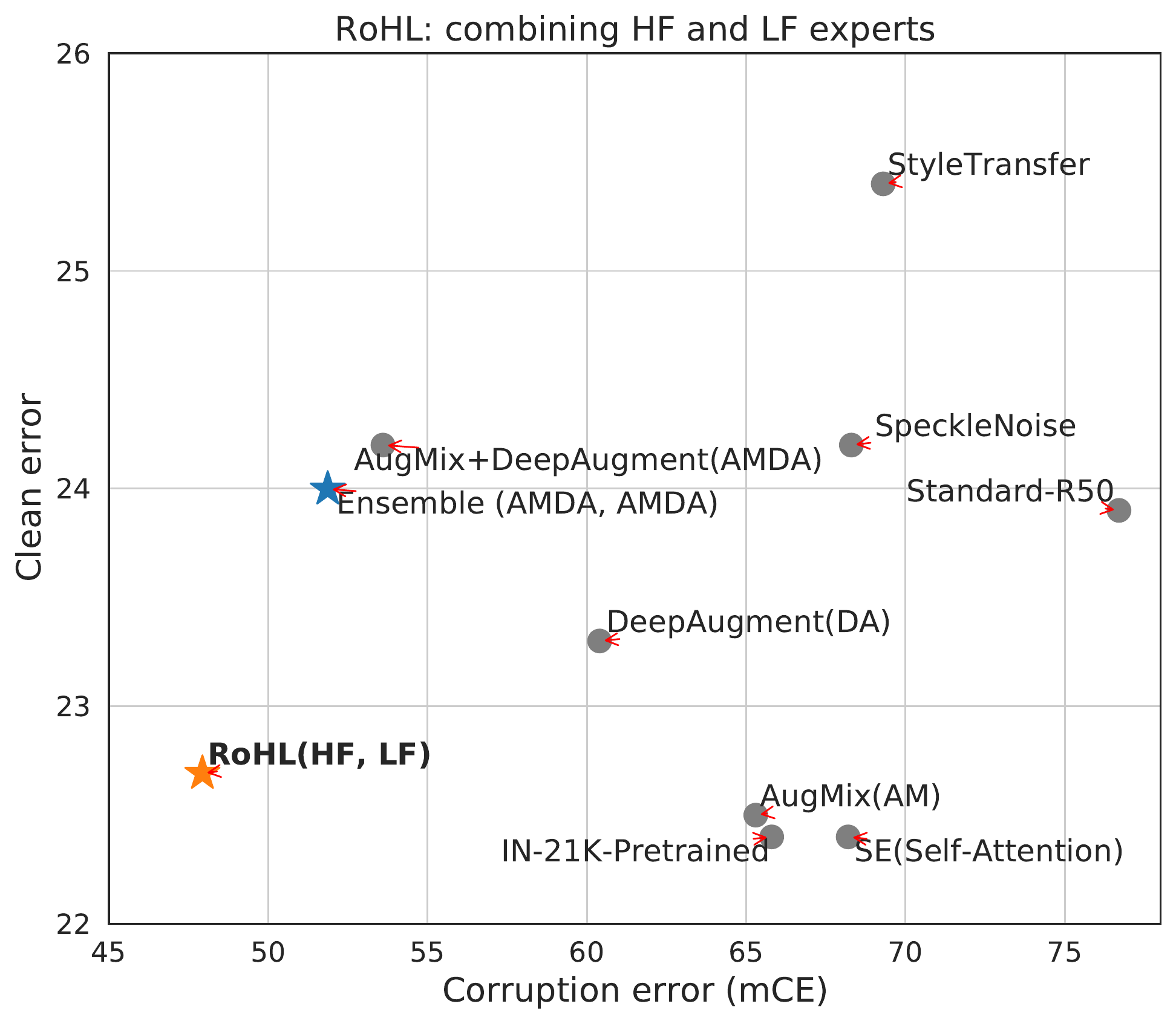}
\caption{ Improving clean and corruption errors. Each item shows the error of a model on ImageNet  (y-axis) and on ImageNet-C (x-axis). All models use a ResNet50 backbone. Orange: The proposed \textbf{RoHL} approach -- \textbf{Ro}bust mixture  of a \textbf{H}F (high-frequency) and a \textbf{L}F (low-frequency) expert model. Blue: An ensemble trained with the state-of-the-art approach AugMix + DeepAugment.  Gray: Other approaches.}
\label{fig:teaser}
\end{center}
\end{figure}

Robustness to distribution shift is possibly the core challenge in deep learning. CNNs show strong performance when training and test set samples are independent and identically distributed (\iid). This led to strong claims of obtaining superhuman performance on the challenging ImageNet dataset. However, such claims have somewhat diminished as the community, driven by practical applications, started testing on out-of-distribution (\ood) test sets. Unlike human vision, CNNs are affected even by small perturbations in the input. Simply adding random noise to the ImageNet test set is sufficient to almost triple the classification error~\cite{hendrycks2019robustness}.

Why does performance drop so severely under distribution shift? One explanation is that models rely on spurious, unstable correlations present in the \iid training and test dataset to obtain low training and test errors. When, due to distribution shift, these unstable correlations are missing, performance drops severely. Although there has been substantial prior work~\cite{geirhos2018generalisation,hendrycks2019robustness,recht-imagenetv2,taori2020measuring,yin2019fourier} investigating this problem, it is far from being fully understood, let alone solved. The most successful remedies to-date are well-chosen data augmentation schemes~\cite{autoaugment,hendrycks2020many,augmix,rusak2020simple,styletransferaug} and adversarial training~\cite{ford2019adversarial,rusak2020simple,wong2020fast}. Geirhos \etal~\cite{geirhos2018generalisation} proposed the \emph{texture hypothesis}, where they show that classification models learn feature representations biased towards textures. Many of these texture features are unstable and get destroyed, for example, due to weather effects or digital corruptions.

The \emph{texture hypothesis} can also be regarded  from a Fourier perspective~\cite{yin2019fourier}. Yin~\etal~\cite{yin2019fourier} showed that models achieve reasonable performance ($\sim$60\% accuracy) on the \iid test set of ImageNet even with strong low or high pass filtering applied to the input images during training and testing.

This indicates the existence of many input-output correlations in  low-frequency and high-frequency domains. They also showed that the performance degradation on corrupted data varies across the frequency spectrum. For instance, standard models trained on clean images are inherently biased to be more robust towards low-frequency corruptions compared to high-frequency ones. It might seem that such biases can be easily fixed with data augmentation. However, data augmentation comes with robustness trade-offs, i.e., many transformations improve performance on some types of corruptions but reduce performance on clean images.  In realistic scenarios, the dominant fraction of data is typically clean and not corrupted. Therefore, clean performance must not be ignored.

To avoid such trade-offs, we propose  \textbf{RoHL} --- \textbf{Ro}bust mixture of a \textbf{H}F (high-frequency) and a \textbf{L}F (low-frequency) expert model. To build the HF expert model, we apply TV minimization~\cite{becht_restoration} on the activations of the first convolutional layer, as well as generic augmentations that affect high-frequency components in the image. The HF expert is robust to high-frequency corruptions whereas the LF expert, based on plain contrast augmentation, is robust to low-frequency corruptions. We show that having such complementary models improves performance both on corrupted and clean images. Also compared to a standard two-member ensemble it adds robustness at no additional cost. An overview of its effectiveness is shown in \figref{fig:teaser}.

In summary, we make two contributions:
\textbf{(1)}~We propose a new regularization scheme that enforces convolutional feature maps to have a low total variation (TV). We show that this boosts high-frequency robustness and is complementary to other high-frequency augmentation operations.
\textbf{(2)}~We introduce the idea of mixing two experts that specialize in high-frequency and low-frequency robustness. We show that this mixture is complementary to diverse data augmentation, such as AugMix~\cite{augmix} and DeepAugment~\cite{hendrycks2020many}.

\section{Related work}

\noindent 
\textbf{Lack of robustness under distribution shift.} Geirhos \etal~\cite{geirhos2018generalisation} and Vasiljevic~\etal~\cite{vasiljevic2016examining}  showed that models trained against certain distortions often fail to generalize to unseen distortions. Hendryks \etal~\cite{hendrycks2019robustness} proposed a synthetic benchmark (ImageNet-C) to study robustness against diverse image distortions.  Recht \etal~\cite{recht-imagenetv2} recreated a new "ImageNetV2" validation set to benchmark naturally occurring domain shift over time and observed larger performance drops.
Recent works evaluated performance under distribution shifts for other vision tasks such as object detection~\cite{objdetrob} and  segmentation~\cite{Kamann2019}, with similar conclusions.
\noindent 
\textbf{Vulnerability to adversarial perturbations.} Adversarial perturbations~\cite{carlini2017adversarial,szegedy_intriguing} are crafted noise signals designed to maximally confuse a model. These perturbations are categorized into white-box attacks~\cite{dong2018boosting,madry2017towards, moosavi2016deepfool, papernot2016limitations, szegedy_intriguing}, where the attacker has accessibility to model weights and gradients and black-box attacks~\cite{brendel2017decision, chen2017zoo, dong2019efficient}, where the attacker can only query the model. Here, we focus on robustness to common corruptions, which are encountered in practice even without an adversary.

\noindent 
\textbf{Improving robustness.} Methods for improving robustness can be broadly grouped into two primary categories: \textbf{a)}  using larger models and datasets ~\cite{hendrycks2020many, orhan2019robustness, xie2020self} \textbf{b)} using data augmentation~\cite{styletransferaug,hendrycks2020many,augmix,rusak2020simple}. Hendryks~\etal~\cite{hendrycks2019using,hendrycks2020many} showed that pre-training on large datasets such as ImageNet-21k improves robustness. Xie~\etal~\cite{xie2020self} trained large models on ImageNet and YFCC100M~\cite{yfcc} in a semi-supervised manner to obtain improved \iid and \ood performance. Taori~\etal~\cite{taori2020measuring} claimed that larger datasets improve performance on \ood data, but are far from closing the performance gap.  An effective measure to improve \ood performance is data augmentation. Ford \etal~\cite{ford2019adversarial} observed that augmentation techniques such as Gaussian or adversarial noise bias the model to be robust against certain corruption types, while degrading on others. Yin~\etal ~\cite{yin2019fourier} showed that these trade-offs can be better understood by looking at the Fourier statistics of the different corruption types. Geihos~\etal~\cite{styletransferaug} showed that using stylized images for training increases shape-bias and thus, improves robustness. Rusak \etal~\cite{rusak2020simple} studied noise corruptions and established a strong baseline on ImageNet-C. Hendryks~\etal~\cite{hendrycks2020many, augmix} showed that diverse data augmentation can obtain strong results on the ImageNet-C benchmark. Recently, Schneider~\etal~\cite{schneider2020better} showed that performance can be further improved by adapting batch-norm statistics at test-time.

\section{Effect of data augmentation on robustness}
\subsection{Robustness trade-offs of data augmentation}

\noindent
\textbf{High frequency robustness.} It has been shown that models trained with Gaussian noise or adversarial  training  exhibit improved  resilience  to  corruptions  that  affect the high  frequencies of the signal~\cite{yin2019fourier}. Such corruptions include different noise corruptions like Gaussian or salt-and-pepper noise. Also corruptions that include blur affect the high-frequency components, as they diminish high-frequency image features such as edges. Data augmentation with operations that act on the high-frequencies make the trained model to rely less on high-frequency features and have been shown to improve  robustness to corruptions concentrated in the high-frequency spectrum considerably. However, as they remove high-frequency features from the model, they also reduce performance on clean images considerably.

\noindent
\textbf{Low frequency robustness.} Achieving robustness to low frequency corruptions, such as fog, haze, contrast, is less obvious compared to high-frequency robustness. Natural images are inherently dominated by the low-frequency components. Yin~\etal ~\cite{yin2019fourier} showed that a data augmentation approach such as randomly perturbing low-frequency Fourier components does not improve low-frequency robustness. The perturbation destroys natural image statistics and even degrades performance on corruptions such as fog. They claimed that no clear trade-off exists for low frequency corruptions. We investigate this further in \secref{sec:aug-bias}.

\subsection{Diverse data augmentation}
\label{sec:diverse-aug}
A way to get around the above trade-offs is the simultaneous application of  diverse data augmentation transformations. AugMix and DeepAugment are two such data augmentation methods, which improve robustness across the frequency spectrum.

\noindent
\textbf{AugMix.} AugMix~\cite{augmix} composes image transformations from a variety of augmentation operations taken from AutoAugment~\cite{autoaugment}. It involves sampling $k$ random sequences of augmentation operations, resulting in $k$ augmented images. These augmented images are then mixed element-wise with randomly sampled weighting factors. A final image is obtained by mixing the augmented image again with the clean version. AugMix models are trained with an additional consistency loss to enforce similar responses for the clean and augmented image embeddings. In particular, the Jensen-Shannon divergence (JSD) among the posterior distributions of the original sample and its augmented variants is minimized.

\noindent
\textbf{DeepAugment.} DeepAugment~\cite{hendrycks2020many} uses encoder-decoder networks trained for image super-resolution  and image compression to generate augmented images. Distorted images are generated by passing an image through these networks but with the weights being perturbed by random transformations.
The distorted images are precomputed before using them for training.

\section{\rohl : combining frequency biased models}

Models trained with different robustness biases are likely to make different errors. We hypothesize that combining models with orthogonal low and high frequency biases should boost performance across the frequency spectrum. We propose \rohl based on this hypothesis and show that it is complementary to diverse data augmentation.

\subsection{Data augmentation targeted for high and low frequencies}
\begin{figure}[t]
\begin{center}
\includegraphics[width=0.75\linewidth]{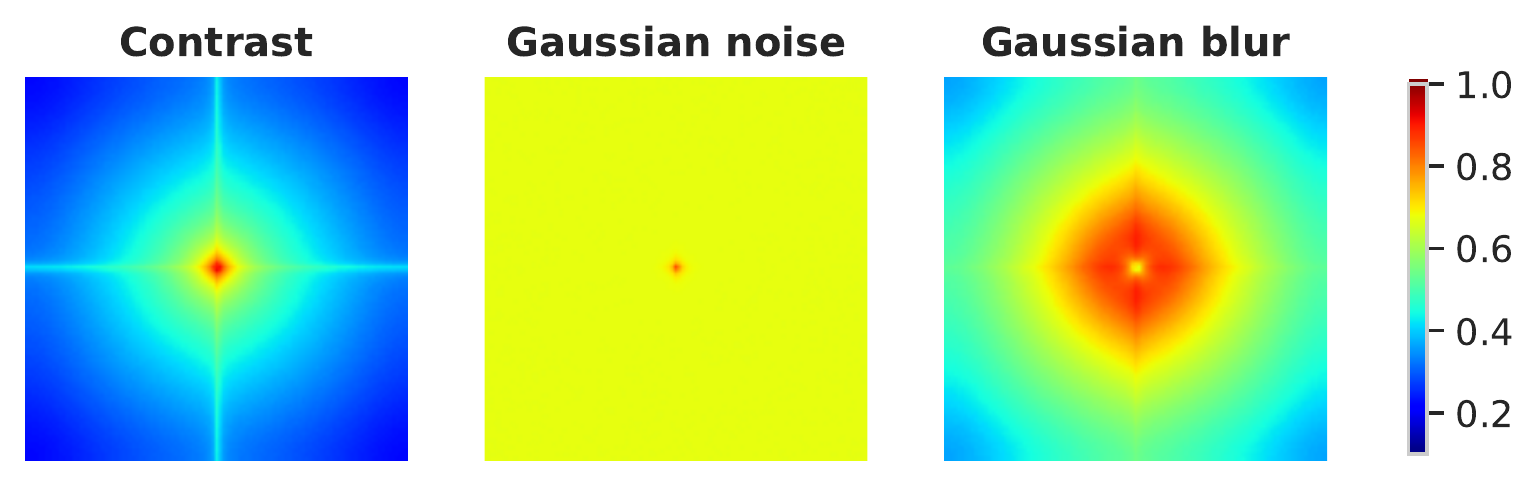}
\caption{Fourier spectrum of three basic corruptions. Low-frequency components are near the center of the spectrum. Left: Contrast augmentation mostly affects the low-frequency components. Middle, Right: Gaussian noise and blur have relatively larger concentrations in high-frequency regions (away from the center). For visualization details; see supplementary (Sec.~1). }

\label{fig:fourier_specturm}
\end{center}
\end{figure}

To cover high-frequency corruptions, we use Gaussian noise and Gaussian blur as generic transformations for data augmentation when training the high-frequency (HF) expert of the ensemble. For added high-frequency robustness we further suggest a new regularization approach when training this expert; see~\secref{sec:tv}.

The second member of the ensemble is optimized for low-frequency (LF) corruptions. We do so by using contrast change as a simple generic augmentation operation that has dominant low-frequency components.

The Fourier spectrum of these simple data augmentation operations is visualized in \figref{fig:fourier_specturm}. Both experts are trained by additionally using diverse data augmentation (we test AugMix and DeepAugment). Implementation details are discussed in the experimental section.

\subsection{Combination of expert predictions}
\label{sec:experts}
The derived expert models for HF and LF robustness are combined and tested on object classification and detection. We combine model predictions by simply averaging predictions of the two member models. We also explored more sophisticated learned merging models. The improvement in performance, however, did not justify the increased complexity over simple averaging (Occam's razor). We denote this combination as \rohl(HF, LF).

\begin{figure}[t]
\begin{center}
\includegraphics[width=0.9\linewidth]{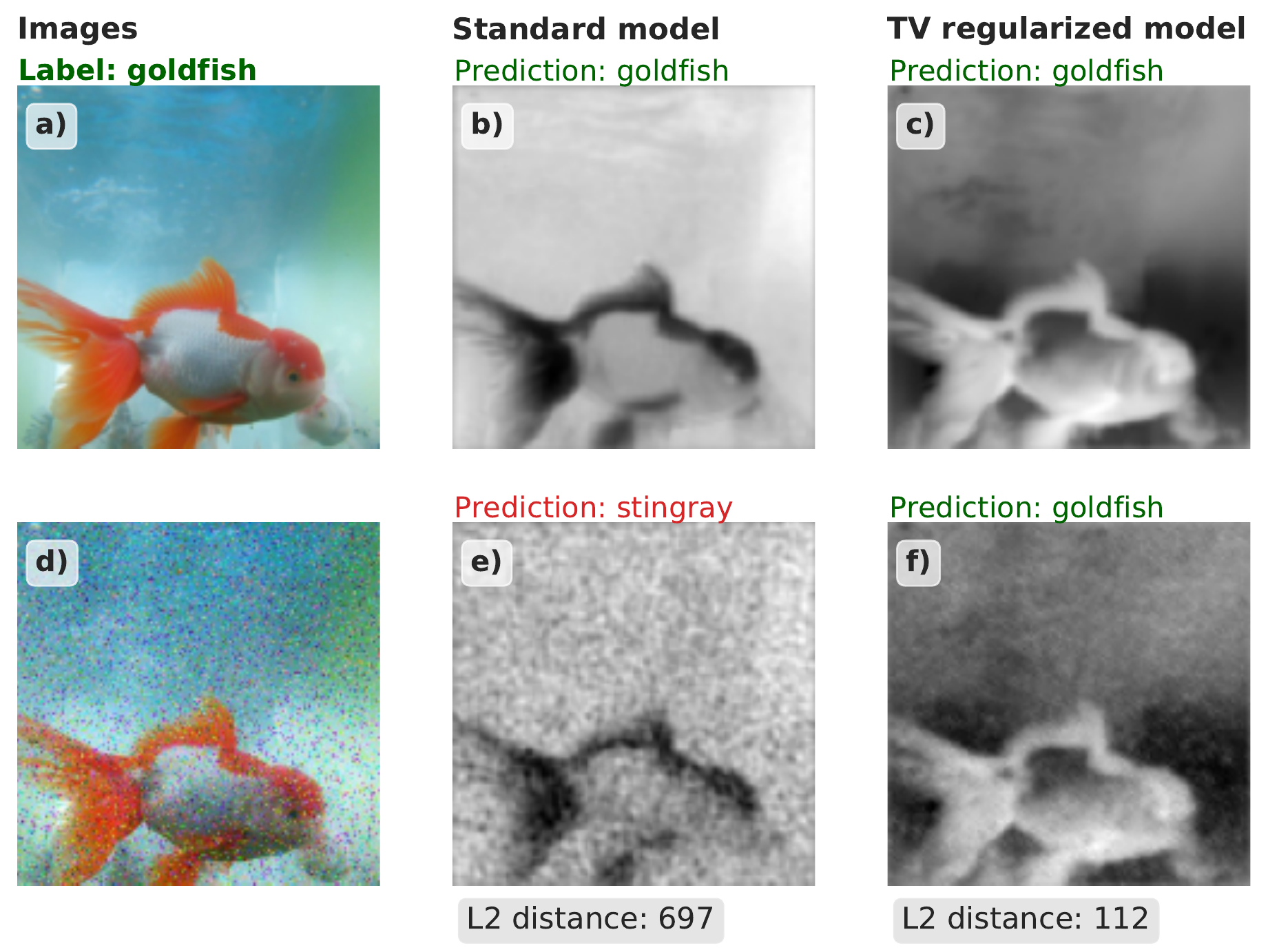}
\caption{Effect of training with TV regularization. a) and d) show a clean and a noisy test image. We compare feature map visualizations of a standard and a TV regularized model. b) and e) show the most active feature map generated after forwarding of a clean and noisy image, respectively. c) and f) show the same for a TV regularized model. Larger activation values have a lighter shade, while smaller values are darker.
We also show the average L2 distance between feature maps from the clean and the noisy \emph{test} images. For a more robust model, the activation statistics should fluctuate less under the influence of noise. We observe that the TV regularized model learns to suppress noise which was unseen during training. We see that f) is much smoother compared to e) and is closer to c). }
\label{fig:feature_maps}
\end{center}

\end{figure}

\subsection{TV minimization on feature maps}
\label{sec:tv}

We improve on the HF expert by introducing a new regularization operation on the early feature maps of the network.
In classical image processing, TV minimization has been widely used for various signal restoration problems~\cite{becht_restoration}. TV minimization is particularly useful for removing oscillations in the signal. Unlike conventional low-pass filtering, TV minimization is a nonlinear operation and is formulated as an optimization problem.

TV minimization could directly filter out noise in the test images, but this requires solving an optimization problem for each test image, which makes the approach slow. Moreover, denoising will also destroy important high-frequency signals and may introduce new artefacts on test images. This can contribute towards additional performance degradation~\cite{hendrycks2019robustness}.

We rather propose to use TV minimization at training time. Instead of applying it to the input images, we apply it to the feature maps of the first conv layer, which processes the input image. As we have discussed, standard CNN models are biased towards using high-frequency information, such as textures. Such a biased model contains filters that fire erratically whenever high-frequency information is present in the input image, resulting in large, noisy activations. This causes downstream layers --- which rely on the first convolutional feature maps --- to behave in unpredictable ways.
We hypothesize that removing spatial outliers (oscillations) in the first conv feature maps will yield more stable representations and, thus, improves robustness to high-frequency corruptions. Since high-frequency signals are picked up best by the first network layer, this is the best placement of the regularizer. We verified this also empirically; see supplementary (Sec. 3).
For continuous functions
$f:\mathbb{R}^{H\times W}\supset\Omega\rightarrow\mathbb{R}$, the TV norm of $f$ is defined as:
\[ \mathcal{L}_{TV}(f) = \int_{\Omega} |\nabla f| \mathrm{ .}
\]
The feature maps $\mathbf{x}\in \mathbb{R}^{H\times W}$ are on a discrete grid. The finite difference approximation reads:
 \[ \mathcal{L}_{TV}(\mathbf{x}) = \sum_{i,j} |x_{i,j+1} - x_{i,j}| + |x_{i+1,j} - x_{i,j}|  \mathrm{.}
 \]

This loss can be combined with the standard cross entropy loss ($\mathcal{L}_{CE}$) for image classification:
\[ \mathcal{L}(\mathbf{\bar{y}},\mathbf{y},\mathbf{F}) = \mathcal{L}_{CE}(\mathbf{\bar{y}},\mathbf{y}) + \lambda \sum_{c}\mathcal{L}_{TV}(\mathbf{F}_{c}) \]
where $\mathbf{F} \in \mathbb{R}^{C\times H\times W}$
denotes conv feature maps with $C$ channels. $\mathbf{\bar{y}}$ and $\mathbf{y}$ denote the predictions and targets respectively. The factor $\lambda$ controls the regularization strength (larger values will result in smoother feature maps). The effect of training models with TV regularization is shown in \figref{fig:feature_maps}. Models trained with TV regularization yield more consistent feature maps for clean and noisy images. We note that this application of TV regularization is different from  standard TV-based image denoising as the reconstruction loss (the data term) is replaced by  cross entropy loss.

\section{Experiments}

\subsection{Experimental setup}

\subsubsection{Datasets}
\noindent
\textbf{ImageNet \& ImageNet-C.} The \mbox{ImageNet} dataset consists of approximately 1.2 million images categorized into 1000 classes.
To evaluate \iid performance we used the standard clean test set. To evaluate performance under distribution shift we used the \mbox{ImageNet-C} dataset~\cite{hendrycks2019robustness}, a corrupted version of \mbox{ImageNet's}  clean test set. \mbox{ImageNet-C} consists of images distorted with 15 different synthetic corruption types (grouped into noise, blur, weather, and digital corruption). Each corrupted subset has 5 severity levels.

\noindent
\textbf{ImageNet-100 \& ImageNet-C-100.} For quicker experimentation, we ran ablations on a smaller subset of the ImageNet dataset consisting of 100 classes. We refer to this dataset as ImageNet-100. The corrupted version of this dataset is denoted as ImageNet-C-100.

\noindent
\textbf{Datasets with natural corruptions.}
To evaluate on natural corruptions we used BDD100k~\cite{bdd100k} and DAWN~\cite{kenk2020dawn}.
BDD100k consists of driving scenes recorded in varying weather conditions and different times of the day. It is an object detection dataset.
We follow~\cite{objdetrob} to create test splits for different weather conditions: clear, rainy and snowy.
DAWN contains a collection of 1000 images taken from road traffic environments with severe weather corruptions. The samples are divided into four weather conditions: fog, rain, snow, and sandstorm.
DAWN is used for testing only.

\noindent
\textbf{Datasets with other distribution shifts.} For non-corruption based shifts we used ImageNet-R~\cite{hendrycks2020many} and ObjectNet~\cite{objectnet}. ImageNet-R contains images of styles, such as abstract or artistic renditions of object classes. ImageNet-R  contains 30k image renditions for 200 ImageNet classes.
ObjectNet contains 50k images with 313 object classes with 109 classes overlapping with ImageNet. Images contain varying pose and background.

\subsubsection {Implementation details}
\textbf{Evaluation.} Classification models are usually compared using the error computed on the clean test set (\iid). The error metric measures the percentage of misclassification and is computed as: $(100 - \text{Top-1-Accuracy})\%$. Besides the clean error, for corruption datasets, we report the \emph{mean corruption error} (mCE). This involves first computing the unnormalized corruption error (uCE$_{c}$) of a given corruption type ($c$) by averaging across the 5 severity levels. Then, for ImageNet-C-100, we average uCE$_{c}$ for all 15 corruption types to compute mCE. For ImageNet-C, we follow the convention~\cite{hendrycks2019robustness} of normalizing (uCE$_{c}$) with AlexNet's corruption error, before averaging over all corruption types. To evaluate classification performance on natural corruptions, we report errors on different corruption types and their mean. For object detection performance, we use the COCO Average Precision (AP) metric, which averages over IoUs between 50\% and 90\%. On corrupted data we also report mean AP over corruption types and denote it as mAPc.

\noindent
\textbf{Architectures.} Our experiments use ResNet50. For ablation experiments on ImageNet-100, we moved to the smaller ResNet18 architecture. The object detection experiments use FasterRCNN~\cite{faster-rcnn} with ResNet50 as backbone.

\noindent
\textbf{Training.}
We employ AugMix data augmentation together with the JSD consistency loss and the default hyperparameters~\cite{augmix}. For DeepAugment, we use augmented images pre-computed by Hendryks~\etal~\cite{hendrycks2020many}. To train with TV regularization, we use a regularization factor $\lambda\!=\!1e^{-5}$ for all experiments (a sensitivity analysis for $\lambda$ is included in the supplementary, Sec. 3). We finetune models to induce HF and LF robustness biases with data augmentation operations. For object detection with FasterRCNN, we used mmdetection framework's implementation~\cite{mmdetection}. For more detailed training settings see supplementary (Sec.~2).

\subsection{Effect of training with TV regularization}
\label{sec:exp_tv}

\begin{figure}[h]

\begin{center}
\includegraphics[width=0.95\linewidth]{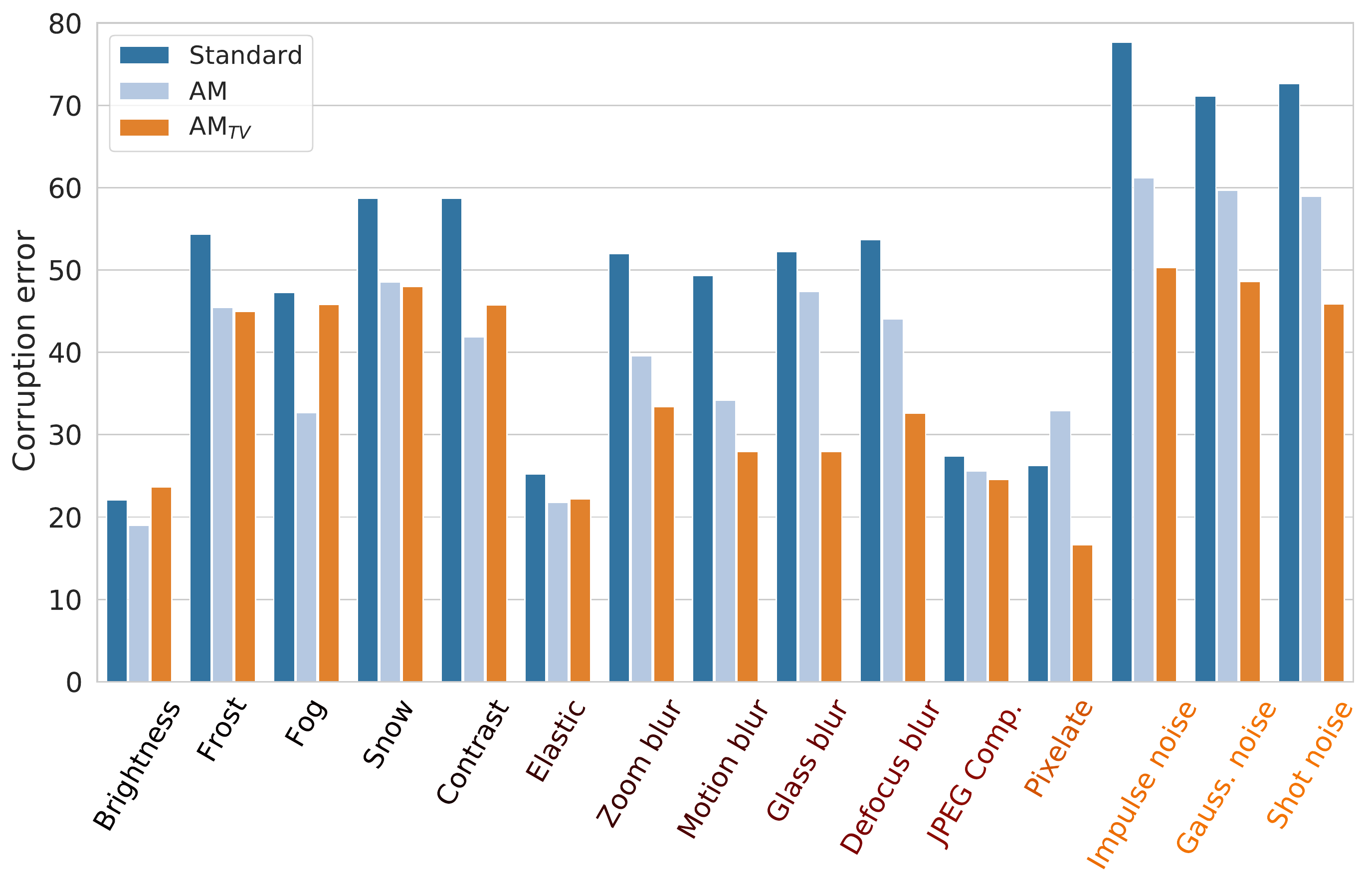}
\caption{Classification error of an \amtv model on different corruption types (ImageNet-C-100). Y-axis: mean error for a given corruption type over all severities. X-axis: corruption types ordered from low to high frequency (indicated by the colour gradient). Ordering is based on the amount of high-frequency content in corruption types; see supplementary (Sec. 1). Standard denotes a baseline model trained on  natural  images. Models trained with AugMix are generally more robust, and TV regularization complements this with consistently better performance on all high-frequency corruptions, making it an excellent high-frequency expert.
}
\label{fig:amtv_perf}
\end{center}
\vspace{-0.1cm}
\end{figure}

We considered the following settings: \textbf{a)} standard baseline model trained on natural images, \textbf{b)} trained with \mbox{AugMix} data augmentation (denoted as \am), \textbf{c)} trained with \mbox{AugMix} data augmentation and TV regularization (denoted as \amtv). \figref{fig:amtv_perf} shows that \textbf{the TV regularized model consistently improves over the standard and the AugMix model on all corruptions that affect high frequencies}. On low-frequency corruptions (Eg: brightness, contrast, fog), TV regularization has a negative effect. Moreover, \tabref{tab:perf-tv} shows that it increases the clean error. This shows that TV regularization induces a \emph{high-frequency robustness} bias, which can be exploited by the proposed high-frequency expert from \secref{sec:experts}.

We also investigated layer-wise application of TV regularization and its impact on the high-frequency robustness. Applying TV regularization on early conv feature maps is crucial for achieving strong high-frequency robustness. Also we evaluated applicability to architectures that do not belong to the ResNet family, namely, DenseNet and MNasNet. Performance gains were similar to ResNet18 with no hyperparameter changes. These additional results are included in the supplement (Sec.~3).

\begin{figure*}[h]
\begin{subfigure}{\columnwidth}
\centering
\includegraphics[width=0.99\linewidth]{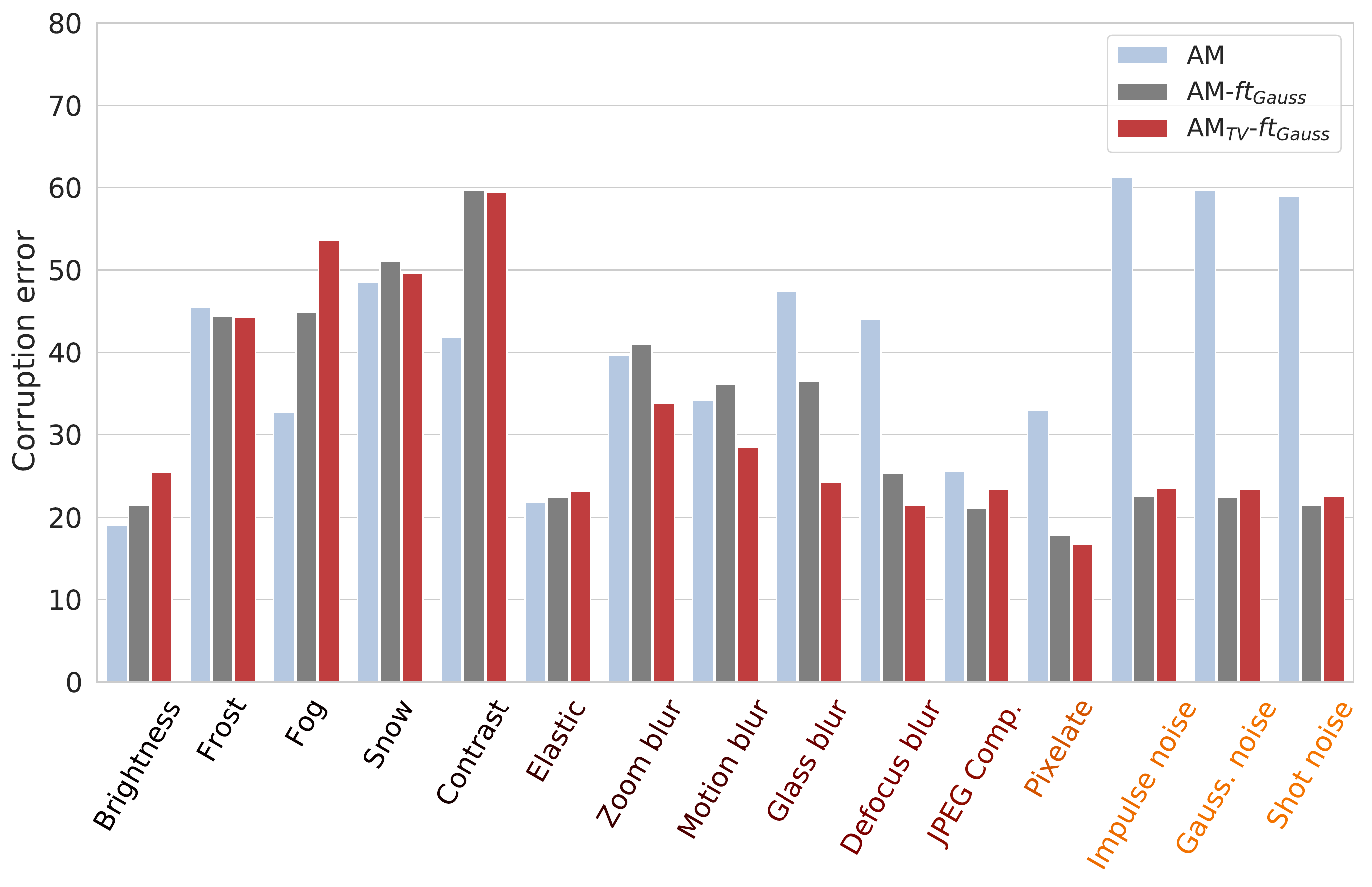}
\caption{High frequency robustness bias}
\label{fig:corr_hf}
\end{subfigure}
\begin{subfigure}{\columnwidth}
\centering
\includegraphics[width=0.99\linewidth]{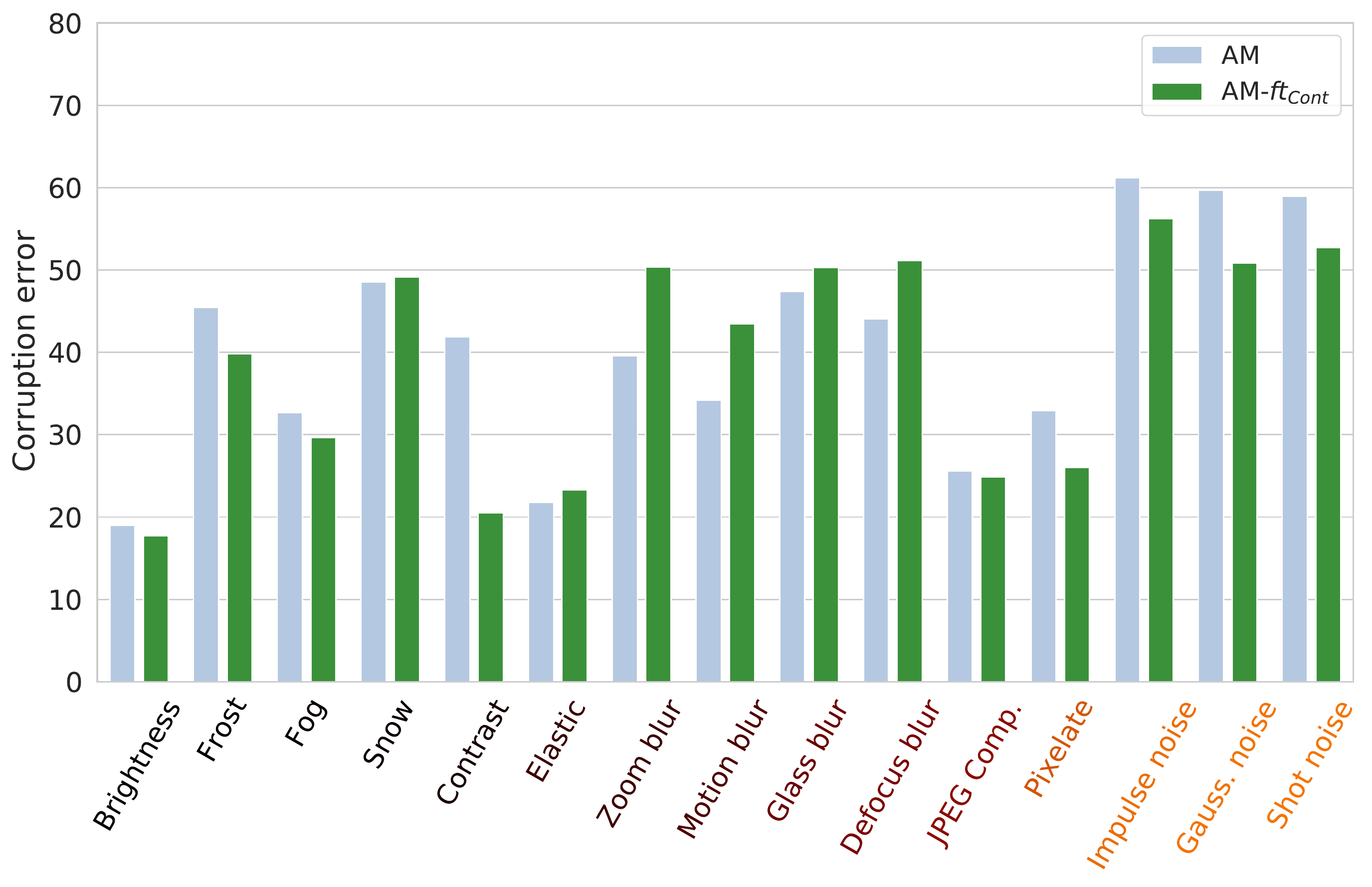}
\caption{Low frequency robustness bias}
\label{fig:corr_lf}
\end{subfigure}
\caption{Robustness bias and its impact on performance across corruption types. Figures \ref{fig:corr_hf} \& \ref{fig:corr_lf} show corruption errors for  models exhibiting high and low frequency robustness biases, respectively. Y-axis: corruption error for different corruption types (averaged over severity levels). X-axis: corruption types ordered from low to high-frequency. In \figref{fig:corr_hf}, both \amtv and \amtvh are robust to high frequency corruptions. \amtvh shows larger improvements on blur corruptions. \figref{fig:corr_lf} shows that \aml improves on low-frequency corruption types. Surprisingly, it also improves performance on some noise corruptions. Comparing figures \ref{fig:corr_hf} \& \ref{fig:corr_lf}, we see that these models have very different biases. }
\end{figure*}

\begin{table}[t]

\begin{center}
\caption{Classification error of the TV regularized model compared to regular training and training with AugMix (ImageNet-100). Standard: baseline model trained on natural images. TV regularization considerably improves on the corrupted test set, but increases the error on clean images.}
\label{tab:perf-tv}
\resizebox{0.42\columnwidth}{!}{
\begin{tabular}{lcc}
\toprule
 Model &  Clean err. & mCE \\
\midrule
 Standard &       12.2      &      49.9 \\
 \am      &       \bf{11.8} &      40.9 \\
 \amtv    &       14.8      &  \bf{35.9} \\
\bottomrule
\end{tabular}
}
\end{center}
\vspace{-0.9cm}
\end{table}

\subsection{Inducing targeted robustness biases}
\label{sec:aug-bias}

\begin{table}[h]
\begin{center}
\caption{Robustness bias due to data augmentation (results on ImageNet-100). Finetuning with Gaussian noise and Gaussian blur induces a high-frequency robustness bias, whereas using contrast augmentation induces a low-frequency robustness bias.}

\label{tab:aug-bias}

\resizebox{0.58\columnwidth}{!}{
\begin{tabular}{lccc}
\toprule
 Model &  Rob. bias & Clean err. &  mCE \\
\midrule
   \am & -     &  \bf{11.8} &      40.9 \\
  \aml &  LF   &  \bf{11.8} &      39.1 \\
  \amh &  HF     &  13.2 &      32.5 \\
 \amtv &  HF     &  14.8 &      35.9 \\
 \amtvh &  HF     &  16.0 &     \bf{31.5} \\
\bottomrule
\end{tabular}
}
\end{center}
\end{table}

\subsubsection{High frequency robustness}
We have seen previously that TV regularization reduces error on high-frequency corruptions at the cost of a higher error on clean images and low-frequency corruptions. In particular, we observed improved robustness for noise and blur corruptions. We tested to what degree this effect can be achieved by finetuning the AugMix models with Gaussian noise and Gaussian blur augmentation applied to the images. We used additive Gaussian noise sampled from $\mathcal{N}(0, 0.08)$. For Gaussian blur, we used a kernel size of 3. We finetuned both \am and \amtv models with these HF augmentation operations. We denote these models  as \amh and \amtvh.

\tabref{tab:aug-bias} shows that TV regularization combined with HF augmentation operations obtains the best mCE. Although the gap compared to \amh seems small, these gains are more pronounced for blur corruptions (see \figref{fig:corr_hf}). Thus, \textbf{TV regularization has a complementary effect to Gaussian noise and blur augmentation}. As we add more high-frequency robustness bias, performance on clean images and low-frequency corruptions deteriorated.

\subsubsection{Low frequency robustness}
To induce robustness on low-frequency distortions, we finetune with contrast augmentation, which is a simple generic transformation that mainly affects the low-frequency components (see \figref{fig:fourier_specturm}).

Yin \etal \cite{yin2019fourier} evaluated a data augmentation scheme by explicitly adding noise to low-frequency Fourier components, and found that such an approach degrades performance on low-frequency corruption types such as fog --- suggesting that a clear trade-off does not exist. On the contrary, we observe that \textbf{finetuning models with a low-frequency perturbation such as contrast augmentation \emph{does} improve performance on other low-frequency corruptions (fog, frost, brightness). Also it does not degrade the clean error}, as shown in \tabref{tab:aug-bias}. \figref{fig:corr_lf} shows that it also improves performance for certain high-frequency corruptions like noise while degrading it on blur. This suggests that trade-offs are more nuanced compared to high-frequency augmentation operations.

\subsection{Combining frequency biased models}
\label{sec:model-combinations}

\begin{table}[h]
\caption{Performance comparison to a standard ensemble (ImageNet-100). Model\textsubscript{1} and Model\textsubscript{2} denote the two members. For a standard ensemble, the two models are independently trained but with similar biases (first two rows). Our results (third and fourth row) show improved performance on corruptions while preserving clean performance.
 }\label{tab:model-combinations}
\centering
\resizebox{0.6\columnwidth}{!}{%
% \begin{tabular}{llccc}
% \toprule
%       &      & \multicolumn{3}{c}{Errors} \\
%               \cmidrule(lr){3-5} 
%               Model\textsubscript{1} & Model\textsubscript{2} &  clean &   mce & avg \\
% \midrule
%           %\am &       11.8 &      40.9 & 26.3 \\
%           \am &  \am &       \bf{10.9} &      39.1 & 25.0 \\
%      \am\textsubscript{\it{gauss, cont}} & \am\textsubscript{\it{gauss, cont}} &       11.4 &      33.2 & 22.3  \\
%  \amh & \aml                    & 11.0 &      32.6 & 21.8 \\
%  \amtvh  & \aml                   &       11.6 &      28.0 & 19.8 \\
%  \hline 
%  \amhh   & \aml                   &    11.4 &      28.4 & 19.9  \\
%  \amtvhh & \aml                   &       11.7 &      25.9 & 18.8  \\
% \bottomrule
% \end{tabular}

\begin{tabular}{llcc}
\toprule
               Model\textsubscript{1} & Model\textsubscript{2} &  Clean err. &   mCE \\ 
               \midrule 
                     \am & \am &       \textbf{10.9} &      39.1 \\
        \am\textsubscript{\it{Gauss, Cont}} &   \am\textsubscript{\it{Gauss, Cont}} &      11.0 &      29.0 \\
   \amh & \aml &       11.4 &      28.4 \\
 \amtvh & \aml &       11.7 &      \textbf{25.9} \\
\bottomrule
\end{tabular}

}
\end{table}

Can we improve on corruption without degrading the clean error? \tabref{tab:aug-bias} shows that biasing models for high-frequency robustness improves the corruption error but degrades the clean error. \aml models retain performance on the clean dataset while improving performance on some corruptions, mostly the low-frequency ones. Since these two models have different frequency biases, it is natural to ask --- can we improve performance by combining them?

Since ensembles generally have a positive effect on classification accuracy, we set up standard ensemble baselines to compare the proposed expert ensemble. The first baseline consists of two \am models. As we have seen that additional augmentation operations improve mCE, we consider a second ensemble, where each \am model is finetuned with all the used augmentation operations (Gaussian noise, blur, and contrast in addition to the default AugMix operations). We denote members of the second ensemble as \am\textsubscript{\it{Gauss, Cont}}.
In these baseline ensembles, the member models have the same biases, as they use the same training pipeline.

\begin{figure}[h]
\begin{center}
\includegraphics[width=0.85\linewidth]{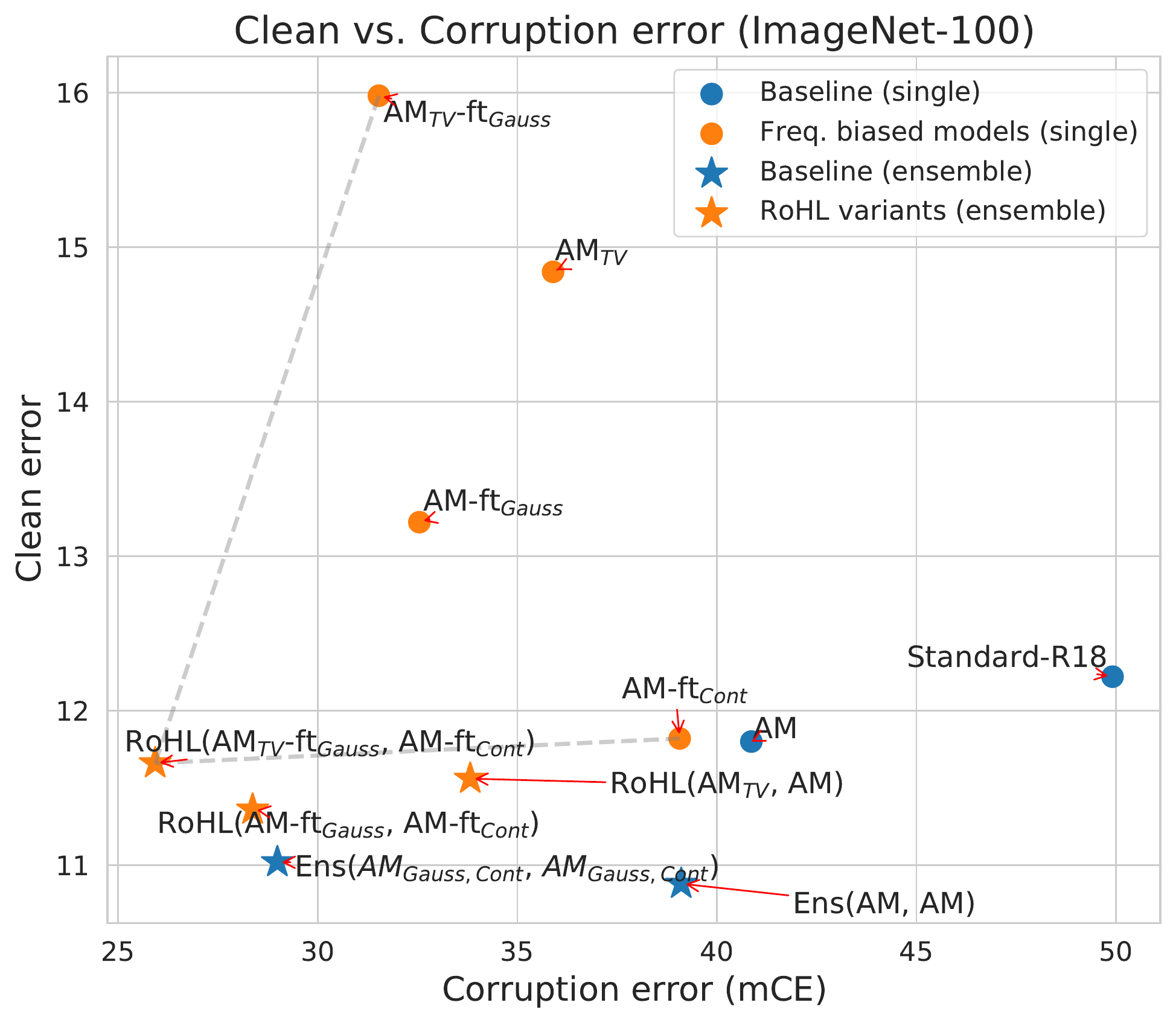}
\caption{Clean vs corruption error on ImageNet-100. Each point represents a model with a certain corruption error (x-axis) and clean error (y-axis). Points closer to the origin indicate a better trade-off between clean and corruption error. Blue: baselines. Orange: variants of RoHL. Dots: single models. Stars: ensembles of two models.
}
\label{fig:model-combinations}
\end{center}
\end{figure}

\tabref{tab:model-combinations} shows that the expert combination \mbox{(\amtvh, \aml)} provides the best clean and corruption error trade-off.  These two models constitute the HF and LF experts for our RoHL approach. It improves the corruption errors by $13.2\%$ points compared to the \am ensemble baseline, while degrading the clean error by only $0.8\%$ points. The trade-off between a low clean error and high robustness to corruptions is best visualized in \figref{fig:model-combinations}, where we plot the clean vs corruption error for various models. \textbf{Combining models with different biases offers a better trade-off than combining models with the same bias.}

\subsection{Scaling to ImageNet}
\label{sec:imagenet_results}
In the previous experiments, we progressively showed training schemes for the HF and LF expert models constituting RoHL. In this section, we verify that the concept carries over to the larger ResNet50 architecture and the full ImageNet dataset. Additionally, we did not just use AugMix for diverse data augmentation, but a combination of AugMix with DeepAugment, a model that was recently suggested by Hendryks~\etal~\cite{hendrycks2020many}.

We first trained a model with TV regularization and AugMix. To train with DeepAugment, we followed Hendryks~\etal~\cite{hendrycks2020many} and finetuned this model with AugMix and DeepAugment (denoted as \amdatv). The high-frequency expert model (denoted as \amdatvh) was obtained by finetuning the \amdatv model with Gaussian noise and blur augmentation. The low-frequency expert was obtained by finetuning the publicly available \amda model with contrast augmentation. We denote this model as \amdal.
\begin{table}[t]
\centering
\caption{Results on ImageNet and ImageNet-C. We compare RoHL to other state-of-the art approaches using a ResNet50 architecture and an ensemble of two AMDA models with already improves the state-of-the-art. RoHL shows the best trade-off between clean error and mCE.}
\label{tab:imagenet-sota}
\resizebox{0.88\columnwidth}{!}{%
% \begin{tabular}{lccc} \\ 
% \toprule
%   & \multicolumn{3}{c}{Errors} \\
%               \cmidrule(lr){2-4} 
%                                              Model & clean &   mce & avg \\
% \midrule
%                                     Standard &       23.9 & 76.7 & 50.3 \\
%                                   IN-21K-Pretrain &       \bf{22.4} & 65.8 & 44.1 \\
%                                 SE (Self-Attention) &      \bf{22.4} & 68.2 & 45.3 \\
%                               CBAM (Self-Attention) &      \bf{22.4} & 70.0 & 46.2 \\
%                               AdversarialTraining &       46.2 & 94.0 & 70.1 \\
%                                       SpeckleNoise &       24.2 & 68.3 & 46.2 \\
%                                      StyleTransfer &       25.4 & 69.3 & 47.3 \\
%                                                 AM &       22.5 & 65.3 & 43.9 \\
%                                                 DA &       23.3 & 60.4 & 41.9 \\
%                                               AMDA &       24.2 & 53.6 & 38.9 \\
%                                       AMDA+AMDA &       24.0 & 51.9 & 37.9 \\
%                                       \hline 
%                         \amtv + \am  &       22.2 & 61.1 & 41.7 \\
%                         \amdatv + \amda  &       23.6 & 49.7 & 36.6 \\
%                         \amdatvh + \amdal &       22.9 & \bf{49.1} & \bf{36.0}  \\
% \bottomrule
% \end{tabular}

\begin{tabular}{clcc}
\toprule
                                             & Model &  Clean err. &  mCE \\
\midrule
                                          & Standard~\cite{he2016deep}  &       23.9 & 76.7 \\
                                          \hline 
\parbox[c]{1mm}{\multirow{10}{*}{\rotatebox[origin=c]{90}{\emph{SOTA approaches}}}} & IN-21K-Pretrained~\cite{hendrycks2020many}  &       22.4 & 65.8 \\
                                & SE (Self-Attention)~\cite{hendrycks2020many} &       22.4 & 68.2 \\
                              & CBAM (Self-Attention)~\cite{hendrycks2020many} &       22.4 & 70.0 \\
                              &  AdversarialTraining~\cite{wong2020fast} &       46.2 & 94.0 \\
                              &          SpeckleNoise~\cite{rusak2020simple} &       24.2 & 68.3 \\
                              &       StyleTransfer~\cite{geirhos2018generalisation} &       25.4 & 69.3 \\
                              &         AugMix (AM)~\cite{augmix} &       22.5 & 65.3 \\
                              &      DeeAugmet (DA)~\cite{hendrycks2020many} &       23.3 & 60.4 \\
                       & AugMix+DeepAugment (AMDA)~\cite{hendrycks2020many} &       24.2 & 53.6 \\
                                              \hline 
    \parbox[c]{1mm}{\multirow{4}{*}{\rotatebox[origin=c]{90}{\emph{Ours}}}} &            Baseline Ensemble (AMDA, AMDA) &       24.0 & 51.9 \\
\cline{2-4}
                       &  \rohl(\amtv, \am)  &       \bf{22.2} & 61.1 \\
                       &  \rohl(\amdatv, \amda) &       23.6 & 49.7 \\
                       &  \rohl(\amdatvh, \amdal) &       22.7 & \bf{47.9} \\
\bottomrule
\end{tabular}

}
\end{table}
\tabref{tab:imagenet-sota} and \figref{fig:teaser} compare our RoHL approach to the state of the art for a ResNet50 model. The standard baseline is a model trained on clean images with random cropping and horizontal flipping. Ensemble (AMDA, AMDA) is a two-member ensemble of the state-of-the-art AMDA model trained with  AugMix and DeepAugment. \textbf{\rohl improves on both the clean and the corrupted error over the previous state-of-the-art (AMDA) and also over its ensemble version}.

\subsection{Results on real image corruptions}
\subsubsection{Object classification}
\begin{table}[h]
\centering
\caption{Object classification performance on natural corruptions. We show errors on various weather corruptions in the DAWN-cls test set. DAWN does not have a un-corrupted test set, hence we show results on the "Clear" test split of BDD100k-cls.}
\label{tab:dawn_results}
\resizebox{\columnwidth}{!}{
\begin{tabular}{l cc cccc}
\toprule
Model  &    Clear  &     & Fog & Rain & Sand & Snow \\
\cline{4-7}
       &    error & mCE &  \multicolumn{4}{c}{errors}    \\
\hline
Standard data augmentation	          &  \cellcolor{gray!15}5.3  & \cellcolor{gray!15}23.5     & 26.3 &	16.1 & 30.3 & 21.5 \\
AMDA                  &  \cellcolor{gray!15}4.9 & \cellcolor{gray!15}16.4    & 19.4 & 10.9 & 21.6 & 13.6\\
Ensemble(AMDA, AMDA)  &  \cellcolor{gray!15}4.9& \cellcolor{gray!15}16.2     & 19.0 & 10.8 & 21.4 & 13.5 \\
\rohl(\amdatvh,\amdal)&  \cellcolor{gray!15}\bf{4.7} & \cellcolor{gray!15}\bf{14.5} & \bf{17.7} &	\bf{10.6} & \bf{19.0} & \bf{10.6} \\
\bottomrule
\end{tabular}
}
\end{table}

BDD100k and DAWN are object detection datasets containing multiple object instances per image and hence cannot be directly used in the classification setting. We extracted object images for each class using 2D bounding box annotations to first transform these datasets to the standard classification setting. The transformed variants are denoted as BDD100k-cls and DAWN-cls.

We finetuned our ResNet50 models (pre-trained on ImageNet) on the  "clear" split of BDD100k-cls. For RoHL, we finetune with the HF and LF biases. We evaluated on corrupted test sets of BDD100k-cls and DAWN-cls.

We observed that weather distortions present in BDD100k are rather benign~\cite{kenk2020dawn,objdetrob}. Thus the corrupted test sets do not impact performance of models trained even with standard data augmentation ($\sim\!\!2\%$ gap between \iid and \ood; see supplementary, Sec.~5). DAWN contains more severe distortions and thus, is more challenging (for examples see supplementary, Sec.~7). \tabref{tab:dawn_results} compares performance of RoHL. \textbf{Compared to the baselines, RoHL performs better on all real corruptions.}

\subsubsection{Object detection}
\begin{table}[h]
\centering
\caption{Object detection performance with different ResNet50 backbones used in FasterRCNN. We report AP scores on the "Clear" split of BDD100k and corrupted test sets in DAWN. Higher AP scores are better. mAPc denotes the mean AP over  corruption types.}
\label{tab:dawn_detection}
\resizebox{\columnwidth}{!}{
\begin{tabular}{l cc cccc}
\toprule
Pretrained Backbone  &  Clear&   & Fog & Rain & Sand & Snow \\
\cline{4-7}
       &    AP & mAPc &  \multicolumn{4}{c}{AP}    \\
\hline
Standard data augmentation	          &  \cellcolor{gray!15}31.3  & \cellcolor{gray!15}24.9    & 21.5 &	25.1 & 24.8 & 21.7 \\
AMDA                                  &  \cellcolor{gray!15}32.4  & \cellcolor{gray!15}27.2    & 24.9 & 26.2 & 27.6 & 24.8\\
Ensemble(AMDA, AMDA)                  &  \cellcolor{gray!15}32.4  & \cellcolor{gray!15}27.2    & \bf{25.4} & \bf{26.2} & 27.6 & 24.2 \\
\rohl(\amdatvh,\amdal)                &  \cellcolor{gray!15}\bf{32.6}  & \cellcolor{gray!15}\bf{28.8}    & 24.9 &	24.9 & \bf{28.1} & \bf{33.4} \\
\bottomrule
\end{tabular}
}
\end{table}

To evaluate on object detection, we used the models finetuned on BDD-100k-cls as backbone in the FasterRCNN architecture. To combine predictions for the baseline ensemble and RoHL, we averaged bounding box predictions and class probabilities (both at the RPN and Fast-RCNN stages~\cite{faster-rcnn}). For implementation details, see the supplementary (Sec.~2). \textbf{\tabref{tab:dawn_detection} shows that RoHL improves over the baselines also in the scope of object detection.}

\subsection{Results on other domain shifts}

To measure performance on distribution shifts other than image corruptions, we evaluated RoHL on ImageNet-R and ObjectNet. Similar to the previous sections, we compare to the two-member ensemble of AMDA models.
On ImageNet-R,  \rohl improves the error by $0.7\%$ points.  On ObjectNet, we obtain an improvement of $1.5\%$ points. Gains for these distribution shifts are marginal. This is to be expected, as object pose changes, for example, are high-level modifications not covered by our approach. See supplementary (Sec. 6) for detailed results.

\subsection{Unsupervised domain adaptation}
We evaluated performance of our models after adaptation using Schneider~\etal 's approach of  updating batch-norm statistics at test time~\cite{schneider2020better}.  Note: this approach is applicable if unlabelled \ood samples of the target distribution are available. \tabref{tab:adaptation} shows results on ImageNet-C and DAWN-cls. \textbf{RoHL's improvements are preserved even after adaptation.}
\begin{table}[h]
\centering
\caption{Results after adaptating BN statistics. Errors with \& without adaption are shown in columns adapt and base. }
\label{tab:adaptation}
\resizebox{0.85\columnwidth}{!}{
\begin{tabular}{l cc cc}
\toprule
Model  &  \multicolumn{2}{c}{ImageNet-C} & \multicolumn{2}{c}{DAWN-cls} \\
       &  \multicolumn{2}{c}{mCE} & \multicolumn{2}{c}{mCE} \\
                       & base  & \cellcolor{gray!15} adapt         &   base  & \cellcolor{gray!15} adapt  \\
\hline
Standard               & 76.7 & \cellcolor{gray!15} 62.2           & 23.5      & \cellcolor{gray!15} 16.8\\
AMDA                   & 53.6 &  \cellcolor{gray!15} 45.4          &  16.4     & \cellcolor{gray!15} 13.6 \\
Ensemble(AMDA, AMDA)   & 51.9 &  \cellcolor{gray!15} 44.7          & 16.2      & \cellcolor{gray!15} 13.5 \\
\rohl(\amdatvh,\amdal) & \bf{47.9} & \cellcolor{gray!15} \bf{41.2} & \bf{14.5} & \cellcolor{gray!15} \bf{12.4}  \\
\bottomrule

\end{tabular}
}
\end{table}

\section{Conclusions}

We demonstrated that a mixture of two expert models -- one specializing on corruptions in the high-frequency spectrum of the image and one specializing on the low-frequency ones -- consistently improves the trade-off between a low error on corrupted samples and a low error on regular clean samples. We also showed that this approach adds to the benefits of a regular ensemble of the same size. Moreover, we introduced TV minimization on the first feature map as a new regularization technique, which consistently improves on high-frequency corruptions and is complementary to other measures in this realm. The principle is flexible with regard to the used base model and dataset size. We showed that the gains transfer to real-world corruptions and also apply to object detection. 
\section*{Acknowledgements}
Experiments were mainly run on the Deep Learning Cluster funded by the German Research Foundation (INST 39/1108-1). We also thank Google for donating GCP credits. The research leading to these results was funded by the German Federal Ministry for Science and Education within the project "DeToL -- Deep Topology Learning", and by the German Federal Ministry for Economic Affairs and Energy within the project “KI Delta Learning -- Development of methods and tools for the efficient expansion and transformation of existing AI modules of autonomous vehicles to new domains". It was also funded in part by the French government under management of Agence Nationale de la Recherche as part of the “Investissements davenir” program, reference ANR-19-P3IA-0001 (PRAIRIE 3IA Institute).

{\small
\bibliographystyle{ieee_fullname}
\bibliography{egbib}
}

\newcommand{\beginsupplement}{%
        \setcounter{table}{0}
        \setcounter{figure}{0}
        \setcounter{section}{0}
     }

\newpage
\clearpage

\beginsupplement

\twocolumn[{\centering{ \Huge Supplementary Material}\vspace{3ex} \medbreak \vspace{2ex}}]

%\appendix

\begin{figure*}[h]
\centering
\includegraphics[width=0.75\textwidth]{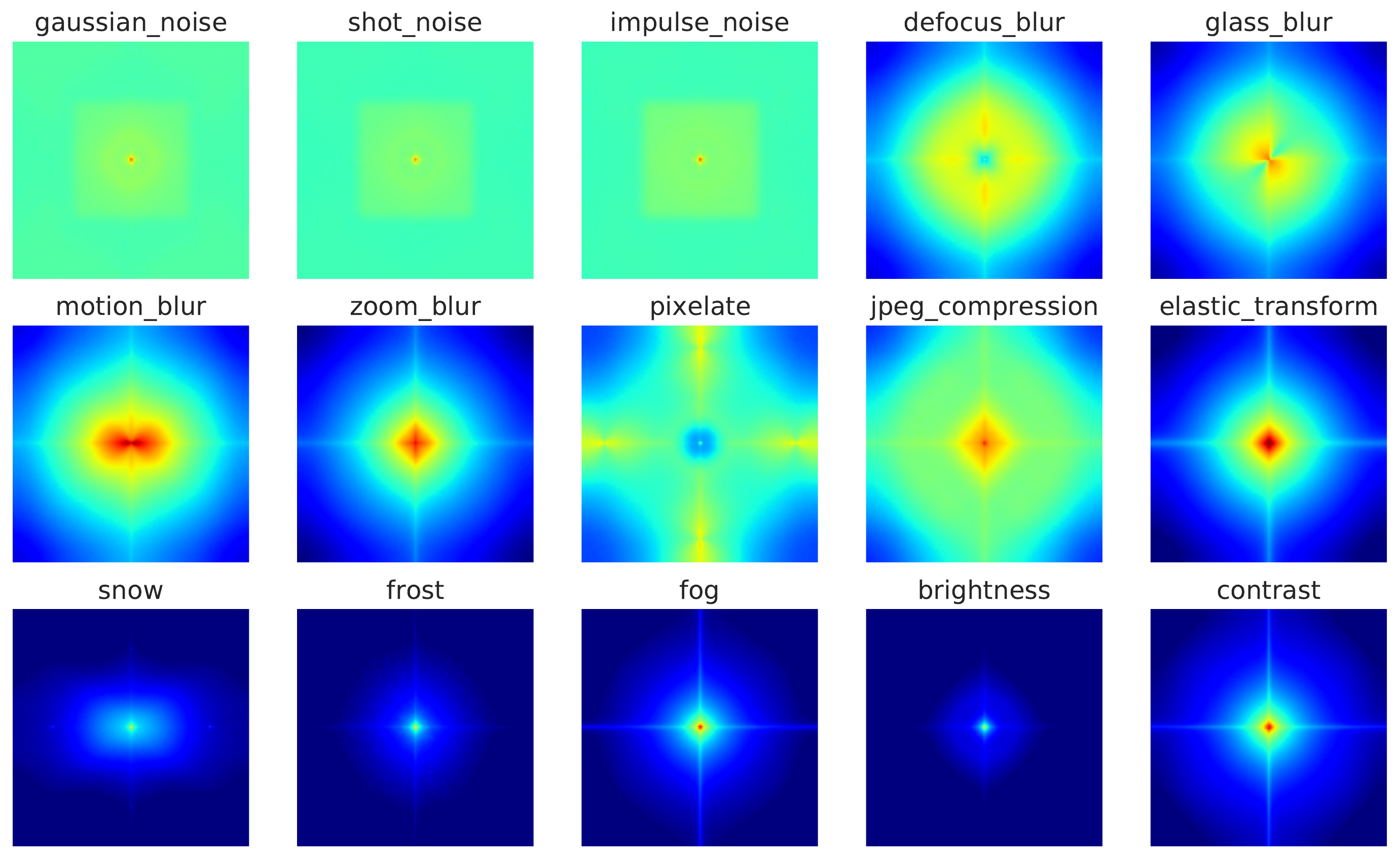}
\caption{ Visualizing Fourier spectrum of different corruption types. Given an image $X$ and a corruption $C$ we plot $\mathbb{E}[|\mathcal{F}(C(X) - X)|]$. $\mathcal{F}$ denotes the 2D discrete Fourier transform. The expectation is computed over $5000$ test images of ImageNet-C for each corruption type. The center shows magnitudes for Fourier components with the lowest frequency. Points away from the center show magnitudes for --- gradually increasing --- higher frequency components. Note: the corrupted images are stored in JPEG format, therefore the visualizations can have some compression artefacts.
}
\label{fig:fourier_specturm_many}
\end{figure*}

\begin{figure}[t]
\centering
\includegraphics[width=\linewidth]{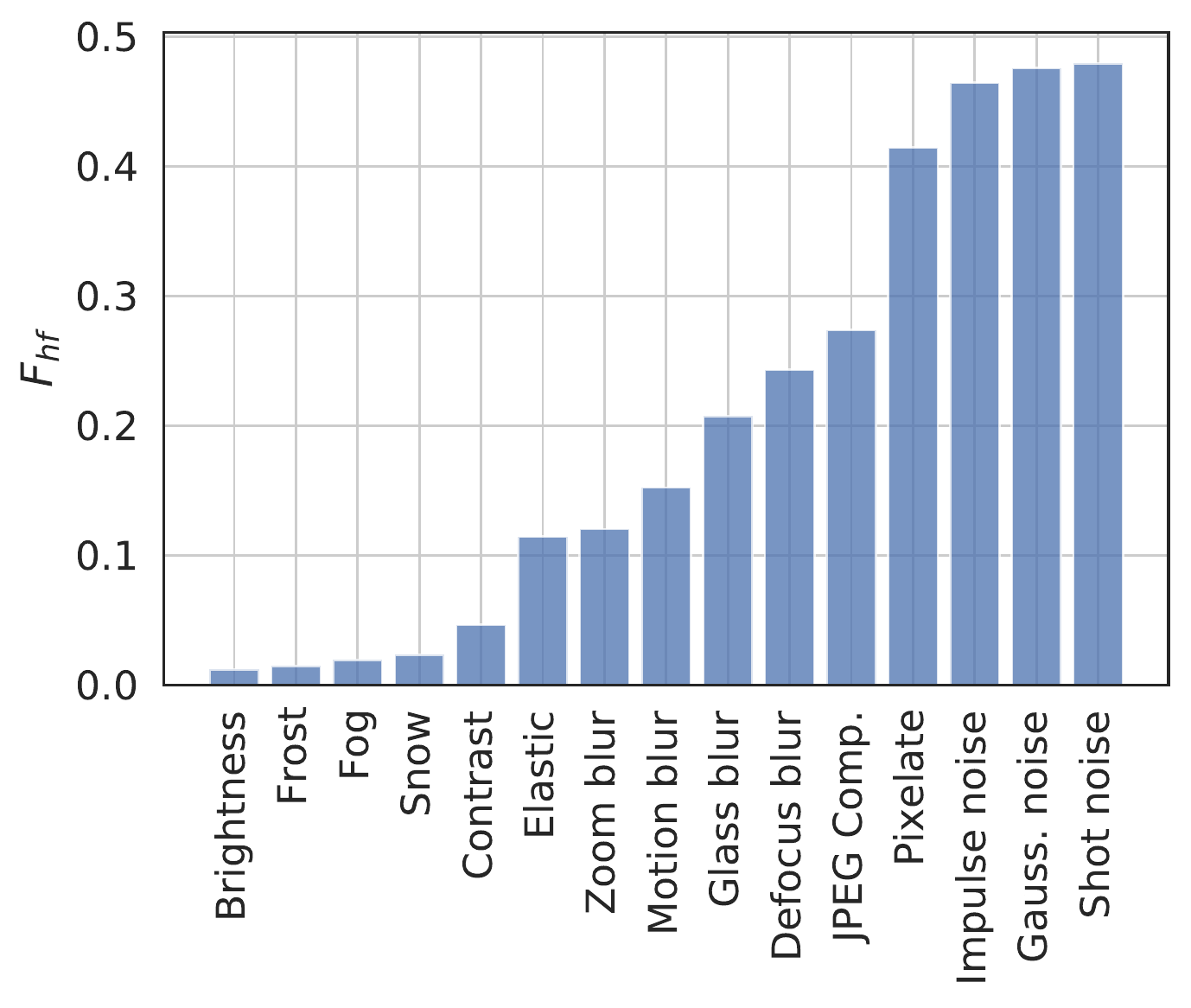}
\caption{Corruption types and their $F_{hf}$ (fraction of high-frequency energy). }
\label{fig:corr-ordering}
\end{figure}

\section{Details on different corruption types}
\label{details}
\subsection{Fourier spectrum visualization}

For visualizing the Fourier spectrum, we always shift low-frequency components to the center of the spectrum. In \figref{fig:fourier_specturm_many}, we visualize the Fourier spectrum of different corruption types in the ImageNet-C test set.
We denote  $\mathcal{F}:\mathbb{R}^{H\times W}\rightarrow\mathbb{C}^{H\times W}$ as  the 2D discrete Fourier transform (DFT).
Given a corruption function \mbox{$C:\mathbb{R}^{H\times W}\rightarrow\mathbb{R}^{H\times W}$} which perturbs a clean image $X$, following Yin~\etal~\cite{yin2019fourier}, we plot
$\mathbb{E}[|\mathcal{F}(C(X) - X)|]$. The quantity $\mathbb{E}[|\mathcal{F}(C(X) - X)|]$ is estimated over $5000$ test images for each corruption type in the first severity level.
We observe that noise and blur corruption types have relatively larger intensities in high-frequency regions (away from the center), compared to corruption types such as fog, frost, brightness, and contrast.

\subsection{Ordering of corruption types}
To visualize induced HF/LF biases, for example, in Figures 4 and 5 of the main paper, we ordered corruption types from low to high frequency. The ordering is done based on the fraction of high frequency energy in the corruption type. Given a clean image $X$ and its corrupted version $C(X)$, the fraction of high frequency energy ($F_{hf}$) of the corruption can be computed as:
\[F_{hf} = \frac{||H(C(X) - X)||^2}{||C(X) - X||^2},\]
where $H(\cdot)$ represents a high-pass filter. We use a circular high-pass filter of size 56.  $\mathbb{E}[F_{hf}]$ is computed over $5000$ images of a given corruption type. \figref{fig:corr-ordering} shows the ordering of corruption types based on $F_{hf}$ values.

\section{Implementation details}
\subsection{Object classification}
\subsubsection{Training}
We used AugMix data augmentation together with the JSD consistency loss~\cite{augmix}. We used the same hyperparameters as~\cite{augmix}.  When training models from scratch we used the default augmentation operations of AugMix. The list of operations is:  autocontrast, equalize, posterize, rotate, solarize, shear, translate. We used the standard $224\times224$ crop size for input images.  For DeepAugment, we used the publicly available augmented images which were pre-computed by Hendryks~\etal~\cite{hendrycks2020many}. We used DeepAugment only for our large scale experiments on ImageNet.

For ImageNet-100, we trained our  ResNet18 models for 75 epochs with AugMix. To train with TV regularization, we used a regularization factor $\lambda\!=\!1e^{-5}$ for all experiments (a sensitivity analysis for $\lambda$s see \figref{fig:lamda-sens}). We observed that these models take longer to converge to a similar training loss as standard AugMix models. Therefore, we train these models for 150 epochs. On single GPUs, we use a batch size of 64 and an initial learning rate of $0.025$ and decayed with the same schedule as~\cite{augmix}.

For ImageNet, we used 8 Nvidia RTX 2080 Ti GPUs to train our ResNet50 models. We train models with AugMix and TV regularization for 330 epochs with a batch size of 256 and an initial learning rate of $0.1$. ResNet50 models trained with AugMix are publicly available, hence we do not re-train these models. For stable distributed training, we follow recommendations of Goyal~\etal~\cite{goyal2017accurate} and perform a warm-up phase by training for 5 epochs. In this phase, the learning rate is linearly increased from $0$ to the initial learning rate of $0.1$.
For training with AugMix and DeepAugment, we follow~\cite{hendrycks2020many}.

For BDD100k-cls, we finetuned our ResNet50 models (pretrained on ImageNet) for $75$ epochs with a batch size of $64$ and initial learning rate of $0.001$.

For all datasets, to induce HF and LF robustness biases we finetuned with the relevant data augmentation operations. The AugMix approach is slightly modified to achieve this. We keep the JSD consistency loss but replace the default list of operations with either HF or LF augmentation operations to induce the required bias.  We finetuned for 15 epochs with an initial learning rate of $0.001$

\subsubsection{Combining predictions} To combine predictions for the baseline ensemble and RoHL, we always use outputs after softmax is applied.
\begin{figure*}[h]
\centering
\includegraphics[width=0.9\linewidth]{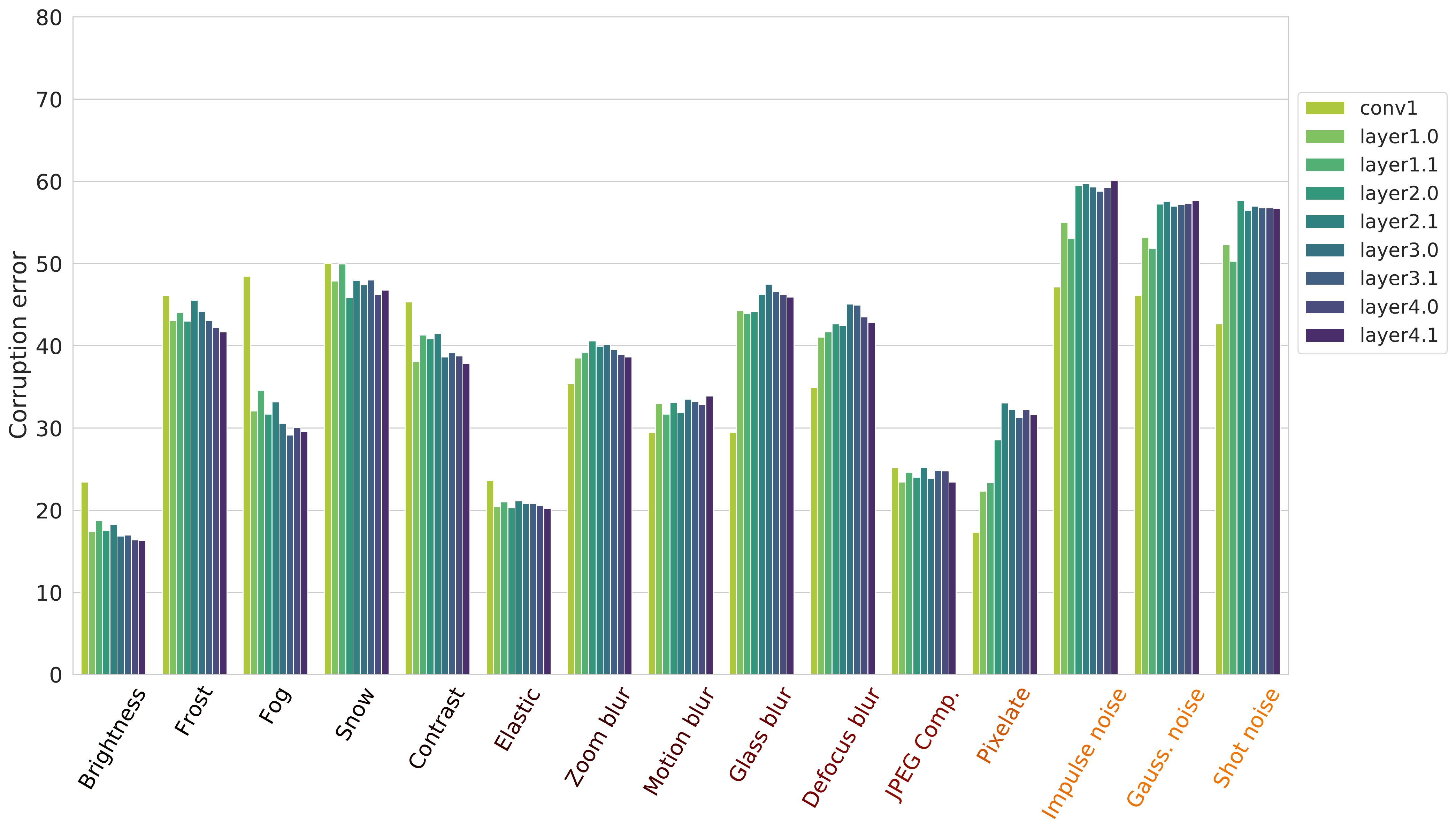}
\caption{Impact of TV regularization applied to different layers (errors on ImageNet-C-100). Y-axis: mean error for a given corruption type over all severities. X-axis: corruption types ordered from low to high frequency. The legend on the right shows models trained with TV regularization applied to a specific layer of the ResNet18 architecture. The layer names are ordered sequentially along the network depth. We observe that applying TV regularization on conv1 --- the first layer that processes the input image --- leads to optimal high-frequency robustness. The effect slowly diminishes as we shift the application of TV regularzation deeper into the network. }
\label{fig:tv-ablation}
\end{figure*}
\subsection{Object detection}
\subsubsection{Training} We use the mmdetection framework~\cite{mmdetection} to train our FasterRCNN architecture. To extract multi-scale feature representations, we used FPN (Feature pyramid networks). We trained using 8 GPUs with the default batch size and initial learning rate. The learning rate is decayed with the "1x" schedule~\cite{mmdetection}. The backbone was initialized with biased ResNet50 models finetuned on BDD100k-cls. We did not induce any further HF/LF biases during FasterRCNN's training.
\subsubsection{Combining predictions}
In addition to class probabilities, object detectors predict bounding box coordinates for each class. FasterRCNN~\cite{faster-rcnn} performs this in two stages. In the first stage, a region proposal  (RPN) head predicts object proposals (rough bounding box estimates irrespective of the object's class) and objectness scores (probability of a proposal containing an object). After non-max suppression, these proposals are refined in the second stage (like Fast-RCNN) where the final bounding box coordinates and class probabilities are predicted. We combine the model predictions also in two stages. In the first stage, each model's object proposals and objectness scores are combined by averaging.  Again, in the second stage, we average class predictions and bounding box predictions estimated by each model.
\subsection{Unsupervised domain adaptation}
For experiments on unsupervised domain adaptation we followed Schneider~\etal~\cite{schneider2020better} to adapt batch normalization statistics. Schneider~\etal have shown that multiple unlabeled examples of the corruptions can be used for unsupervised adaptation. Updating the activation statistics estimated by batch normalization at training time with those of corrupted samples improves performance on ImageNet-C. Before evaluation on a corrupted test set, we used all samples to update the batch normalization statistics. Table 7. of the main paper shows results after adaptation. Note: we are able to preserve state-of-the-art performance on ImageNet-C even after adaptation.

\section{Additional results on TV regularization}
\subsection{Layer-wise application of TV regularization}

As we have discussed in Sec.~4.3 of the main text, standard CNN models are biased towards using high-frequency information, such as textures. Such a biased model contains filters that fire erratically whenever high-frequency information is present in the input image, resulting in large, noisy activations. This causes downstream layers --- which rely on the first convolutional feature maps --- to behave in unpredictable ways.
We hypothesized that removing spatial outliers in the first conv feature maps will yield more stable representations and, thus, improves robustness to high-frequency corruptions. We verify this hypothesis empirically by applying this regularization to different layers of a ResNet18 architecture along the network depth. Results are shown in \figref{fig:tv-ablation}. We observe that applying TV regularization to the first conv layer's activation maps leads to optimal high-frequency robustness.

\subsection{Results on other architectures}

Besides the ResNet family, we evaluated for two additional architectures, MNasNet\_0.75 and DenseNet121. \tabref{tab:tv_arch} shows results on  IN-100 with the same hyperparameters as ResNet. We observe a  significant decrease, similar to ResNet18 for AugMix models (see Table~1 in the main text).

\begin{table}[h]
    \centering
    \caption{Performance of \amtv with other architectures on IN-100. }
    \label{tab:tv_arch}
    \resizebox{0.7\columnwidth}{!}{
    \begin{tabular}{llcc}
    \toprule
         Model   & Arch.           & clean err. & mCE \\
         \hline
          \am    & MNasNet\_0.75   &    11.8    & 45.2 \\
          \amtv  & MNasNet\_0.75   &    11.6    & 39.0 \\
          \hline
          \am    & DenseNet121     &    9.8     & 36.6 \\
          \amtv  & DenseNet121     &    12.7    & 30.4  \\
    \bottomrule
    \end{tabular}}
\end{table}

\subsection{The TV regularization factor $\lambda$}

\begin{figure}[b]
\centering
\includegraphics[width=\linewidth]{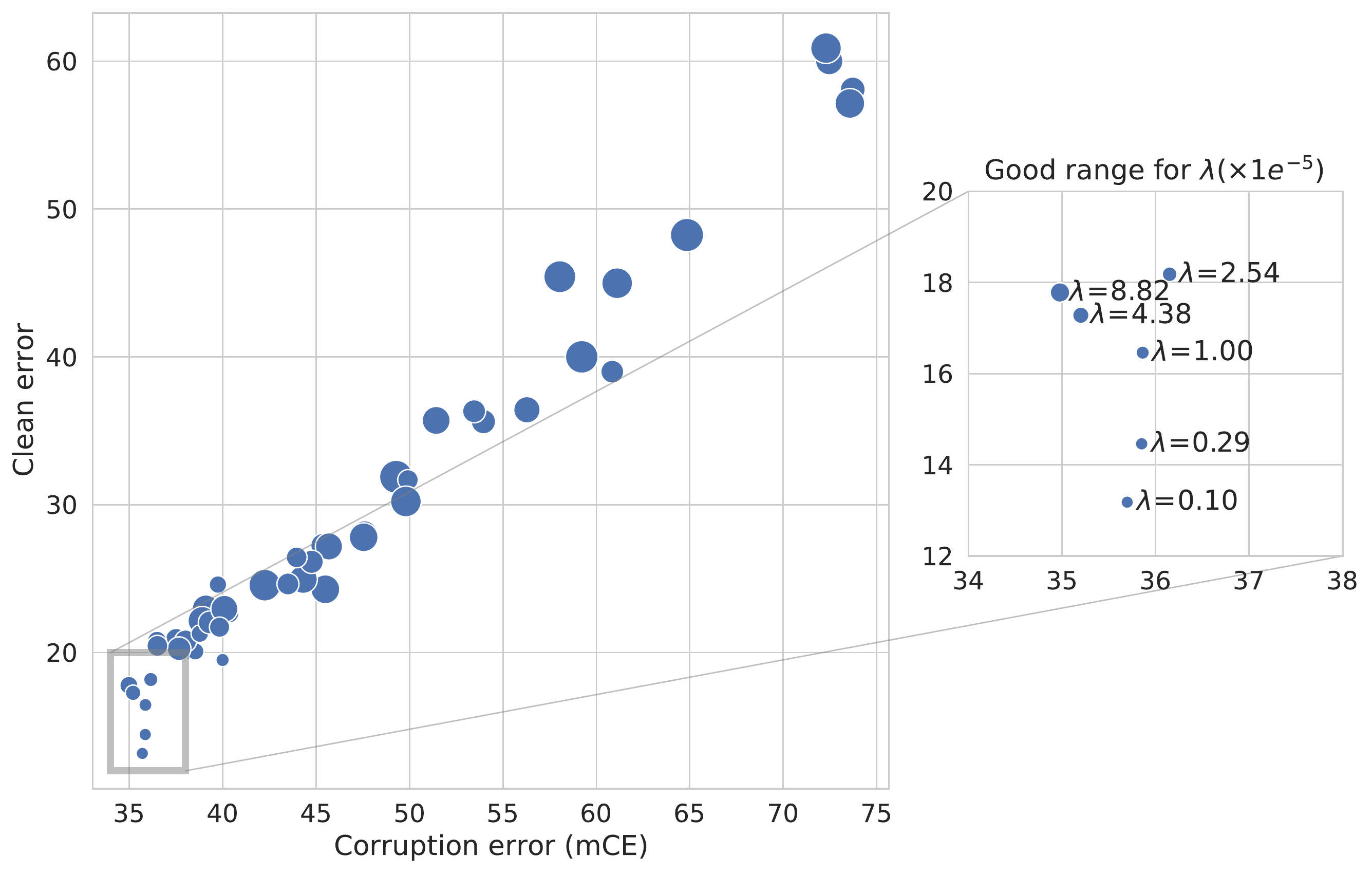}
\caption{ Clean vs corruption error for different values of $\lambda$ (ImageNet-100). Each dot shows the performance of a model trained with a certain $\lambda$. Values of $\lambda$ are sampled uniformly at random from the range: $[1e^{-6}, 1e^{-3}]$. The size of each dot is directly proportional to the sampled value for $\lambda$ (larger dots indicate larger values of $\lambda$). Left: Shows performance of all models. Right: A closer look at models with good clean vs corruption error trade-off. We observe that models trained with smaller regularization factors ($1e^{-6} < \lambda < 9e^{-5}$) perform better.
}
\label{fig:lamda-sens}
\end{figure}

The hyperparameter $\lambda$ controls the strength for the TV regularization term. For all experiments in the main paper, we used a value of $1e^{-5}$. Here we study how different values of $\lambda$ affect the clean and corruption error. To this end, we first sampled $50$ random values for $\lambda$ in the range $[1e^{-6}, 5e^{-4}]$. For each $\lambda$  we trained ResNet18 models on ImageNet-100 with TV regularization. We plot the clean vs corruption error for each model in \figref{fig:lamda-sens}. We observe that models trained with $\lambda \in [1e^{-6}, 9e^{-5}]$ have a good clean vs corruption error trade-off. Larger values of $\lambda$ degrade both clean and corruption errors.
\subsection{Effect of $\lambda$ on feature maps}

\begin{figure*}[h]
\begin{center}
\begin{subfigure}{\linewidth}
        \includegraphics[width=0.95\linewidth]{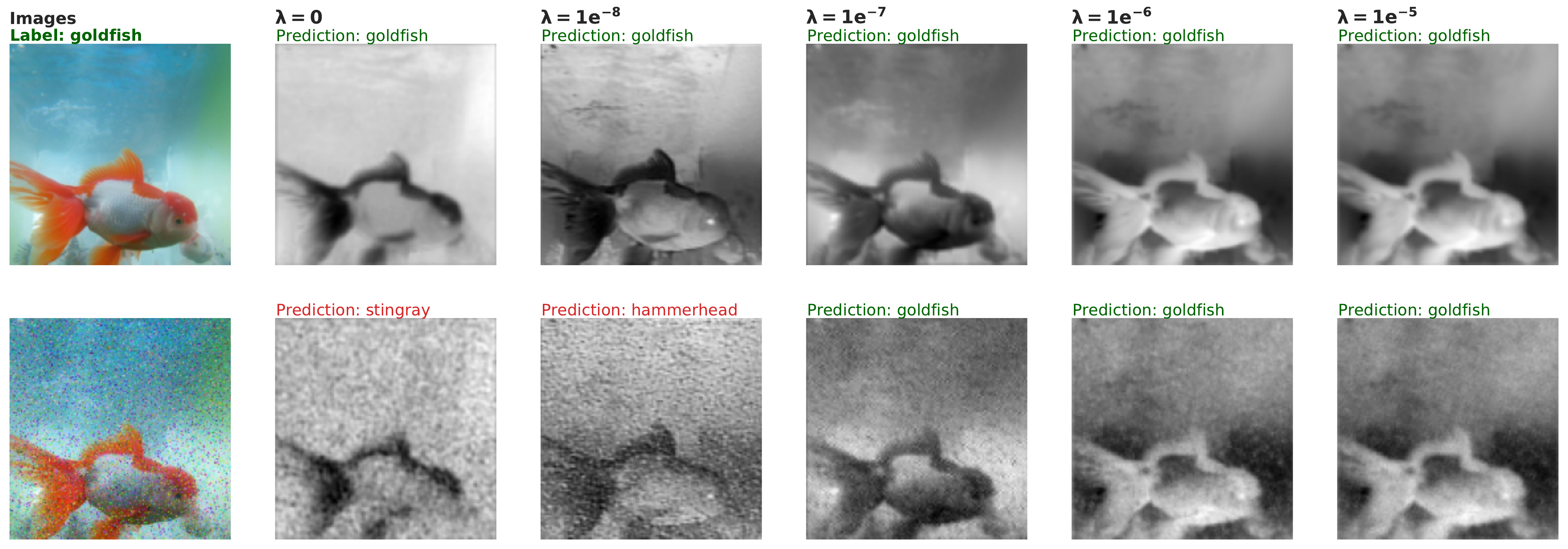}
        \caption{Example: goldfish}
        \label{fig:goldfish}
\end{subfigure}
\begin{subfigure}{\linewidth}
        \includegraphics[width=0.95\linewidth]{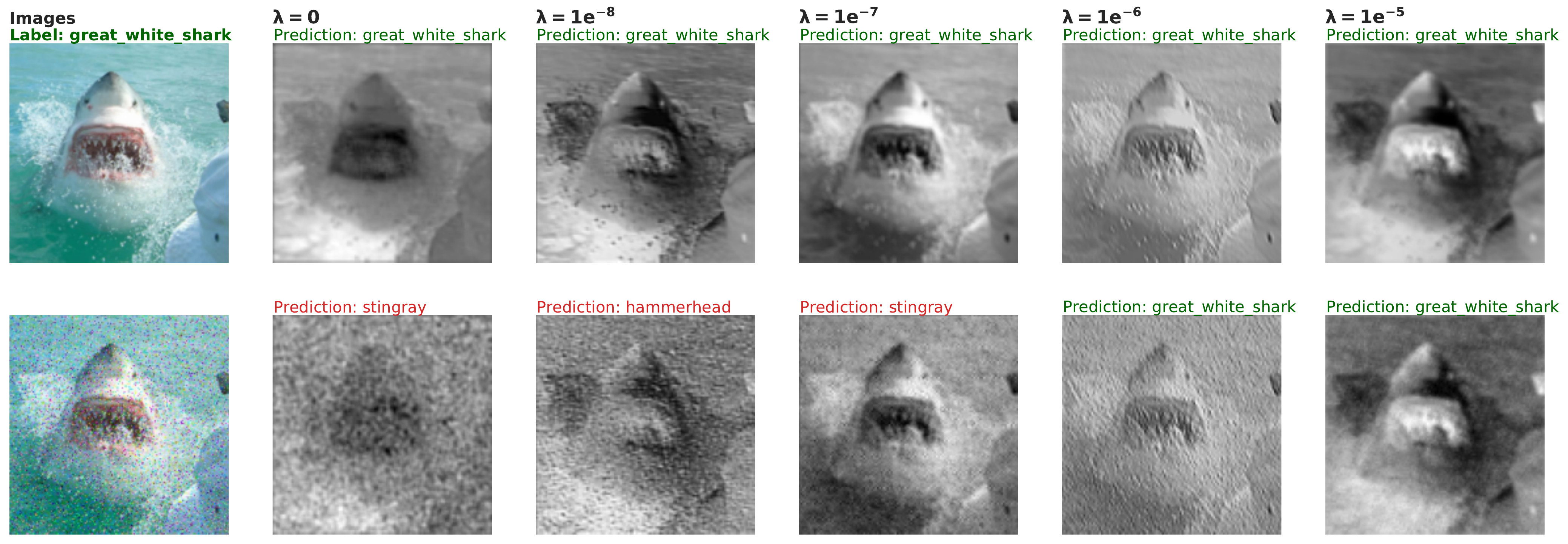}
        \caption{Example: great white shark}
        \label{fig:shark}
\end{subfigure}
\caption{Effect of increasing TV regularization factor ($\lambda$). In \figref{fig:goldfish} and \figref{fig:shark} we visualize two examples to show effect of increasing the TV regularization factor $\lambda$ that is used for training. First column: clean and noisy images. Remaining columns (left to right):  the most active feature map (conv1) generated after forwarding a clean and noisy \emph{test} image to a model trained with a certain $\lambda$ (shown in the column header). Larger activation values have a lighter shade, while smaller values are darker. $\lambda\!=\!0$ means no TV regularization was used. Models with no TV regularization fire erratically for the noisy test image. Increasing $\lambda$ leads to smoother activation maps. With a larger $\lambda$ $(\geq1e^{-4})$, models face convergence issues and performance deteriorates.}
\label{fig:feature_maps_many}
\end{center}
\end{figure*}

In \figref{fig:feature_maps_many} we visualize two examples to show the effect of increasing the TV regularization factor $\lambda$ that is used for training. We observe that as we increase $\lambda$ during training, the most active feature map for the conv1 layer is impacted less by noise at test time. We highlight that these models were not trained with any noise augmentation.

\section{Detailed results on ImageNet}

\begin{figure*}[h]
\begin{subfigure}{\columnwidth}
\centering
\includegraphics[width=0.99\linewidth]{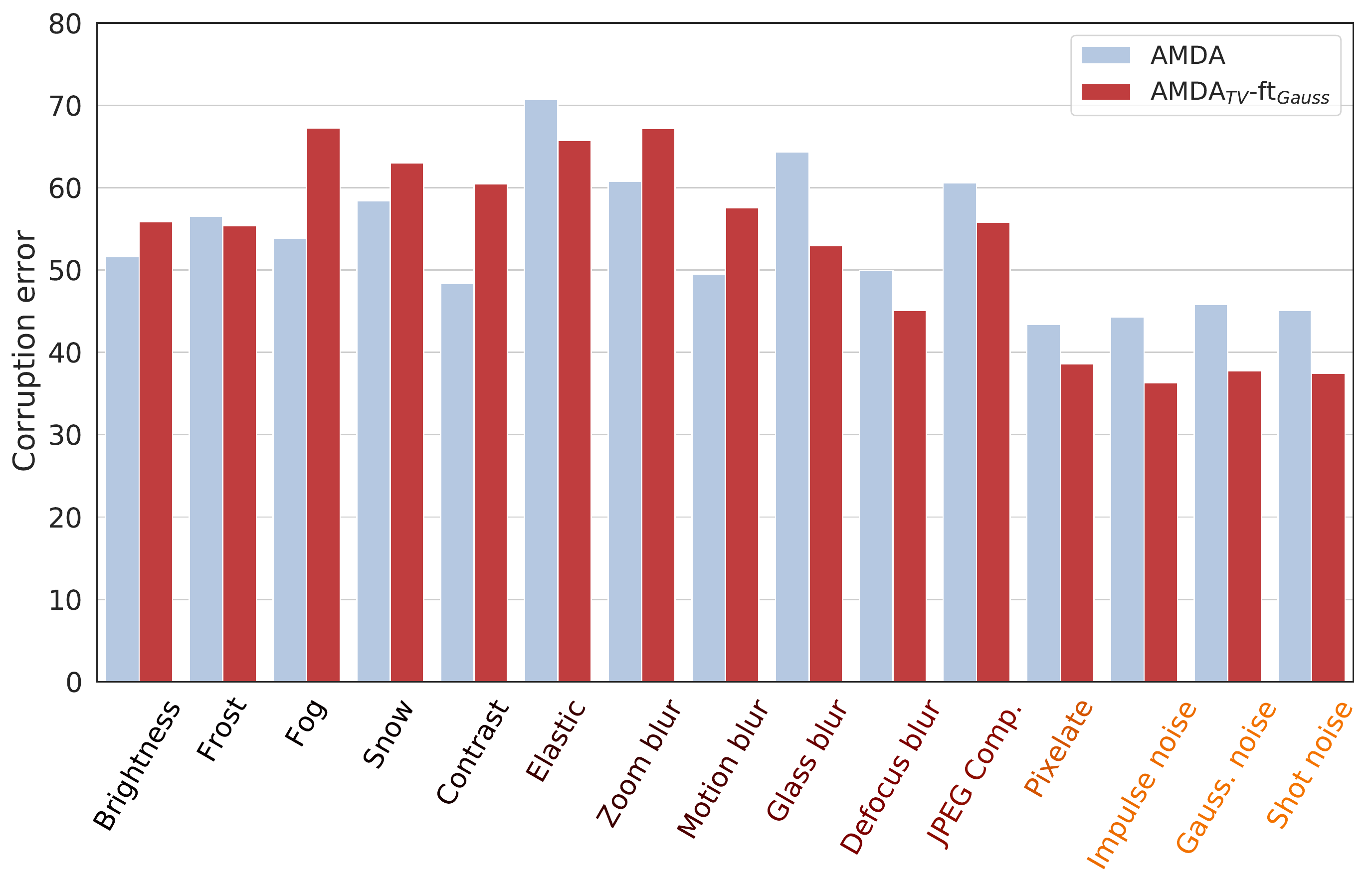}
\caption{High frequency robustness bias}
\label{fig:corr_hf_in}
\end{subfigure}
\begin{subfigure}{\columnwidth}
\centering
\includegraphics[width=0.99\linewidth]{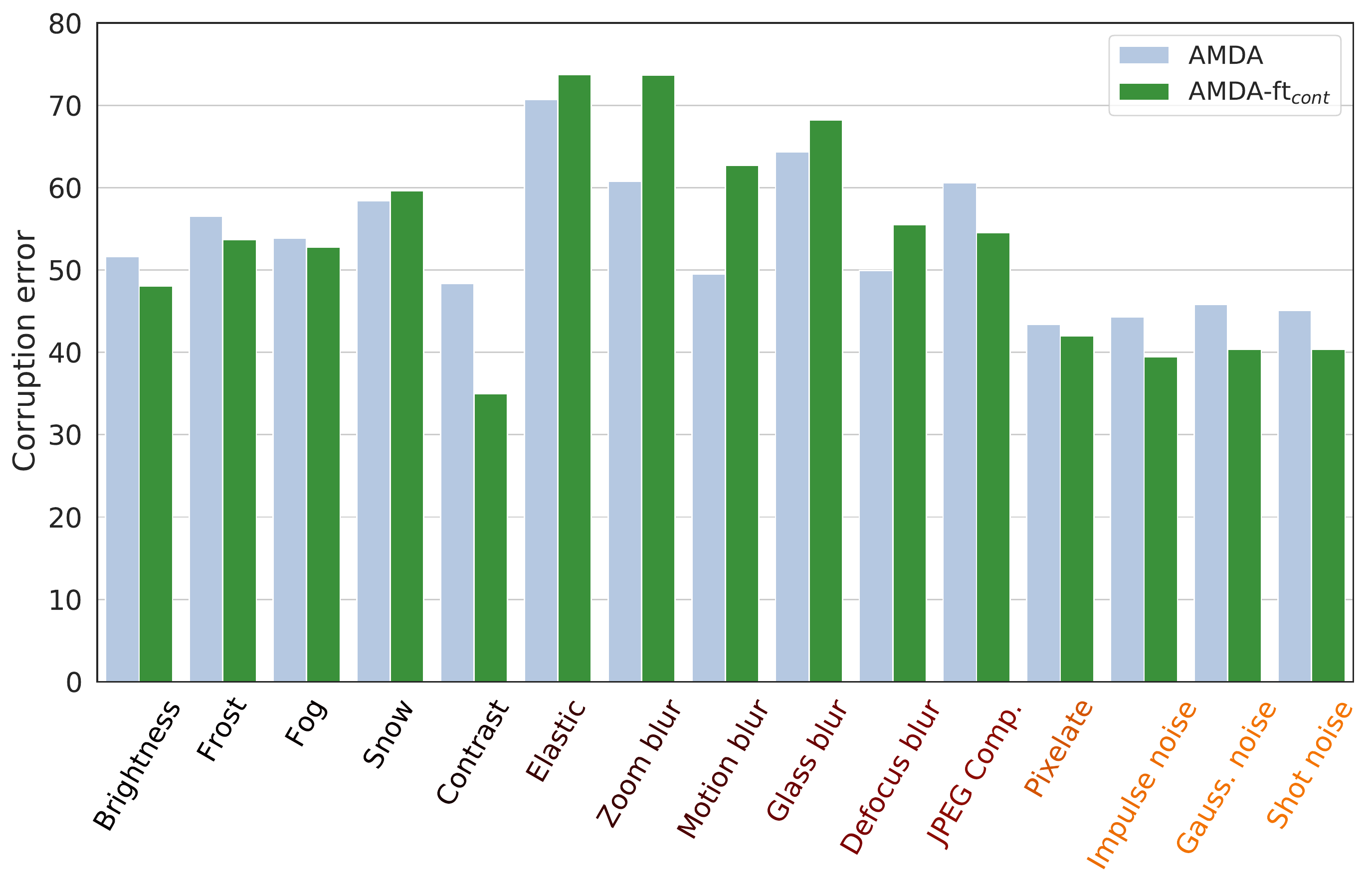}
\caption{Low frequency robustness bias}
\label{fig:corr_lf_in}
\end{subfigure}
\caption{Robustness bias and its impact on performance across corruption types (ImageNet). \figref{fig:corr_hf_in} and \figref{fig:corr_lf_in} show corruption errors for  models exhibiting high and low-frequency robustness biases, respectively. The y-axis shows the corruption error for different corruption types (averaged over severity levels) and the x-axis shows corruption types categorized into high-frequency (red text) and low-frequency (blue text). In \figref{fig:corr_hf_in}, we see that \amdatvh is more robust to high frequency corruptions compared to \amda. \figref{fig:corr_lf_in} shows that \amdal improves on low-frequency corruption types. Surprisingly, it also improves performance on some noise corruptions. Comparing \figref{fig:corr_hf_in} and \figref{fig:corr_lf_in}, we see that these models have very different biases. }
\label{fig:experts}
\end{figure*}

\subsection{Robustness biases of expert models}

\begin{table}[b]
\centering
\caption{Performance of HF and LF experts (ImageNet). We show the clean error and mCE for ResNet50 models trained on ImageNet. High-frequency (HF) expert is \amdatvh. Low-frequency (LF) expert is \amdal.}
\label{tab:single-models}
\resizebox{0.55\columnwidth}{!}{%
\begin{tabular}{lcc}
\toprule
                                           Model &  Clean err &  mCE \\
\midrule
                 AMDA &       24.2 &      53.6 \\
                 \amdal &     \bf{23.4} &      52.8 \\
 \amdatvh &       26.4 &      \bf{52.6} \\
\bottomrule
\end{tabular}
}
\end{table}

We show the clean error and mCE for the high-frequency and low-frequency expert models in \tabref{tab:single-models}.  The high-frequency expert (\amdatvh) was first trained with AugMix and DeepAugment with TV regularization and then finetuned  on Gaussian noise and blur. The low-frequency expert was obtained by finetuning the publicly available \amda model with contrast augmentation.

Although the the results in \tabref{tab:single-models} does not show much difference in terms of mCE, these expert models have very different robustness biases. This is shown in \figref{fig:experts}.
Compared to the baseline \amda, the high-frequency expert \amdatvh improves on most high-frequency corruptions while performing worse on low-frequency corruptions.  \amdal on the other hand improves on most low-frequency corruptions and some high-frequency corruptions (noise). These observations are similar to the small scale ablation experiments on ImageNet-100 in the main paper (Section 5.3). Also we highlight that clean error improves for the low-frequency expert \amdal (see \tabref{tab:single-models}).

\begin{figure}[t]
\begin{center}
\includegraphics[width=0.95\columnwidth]{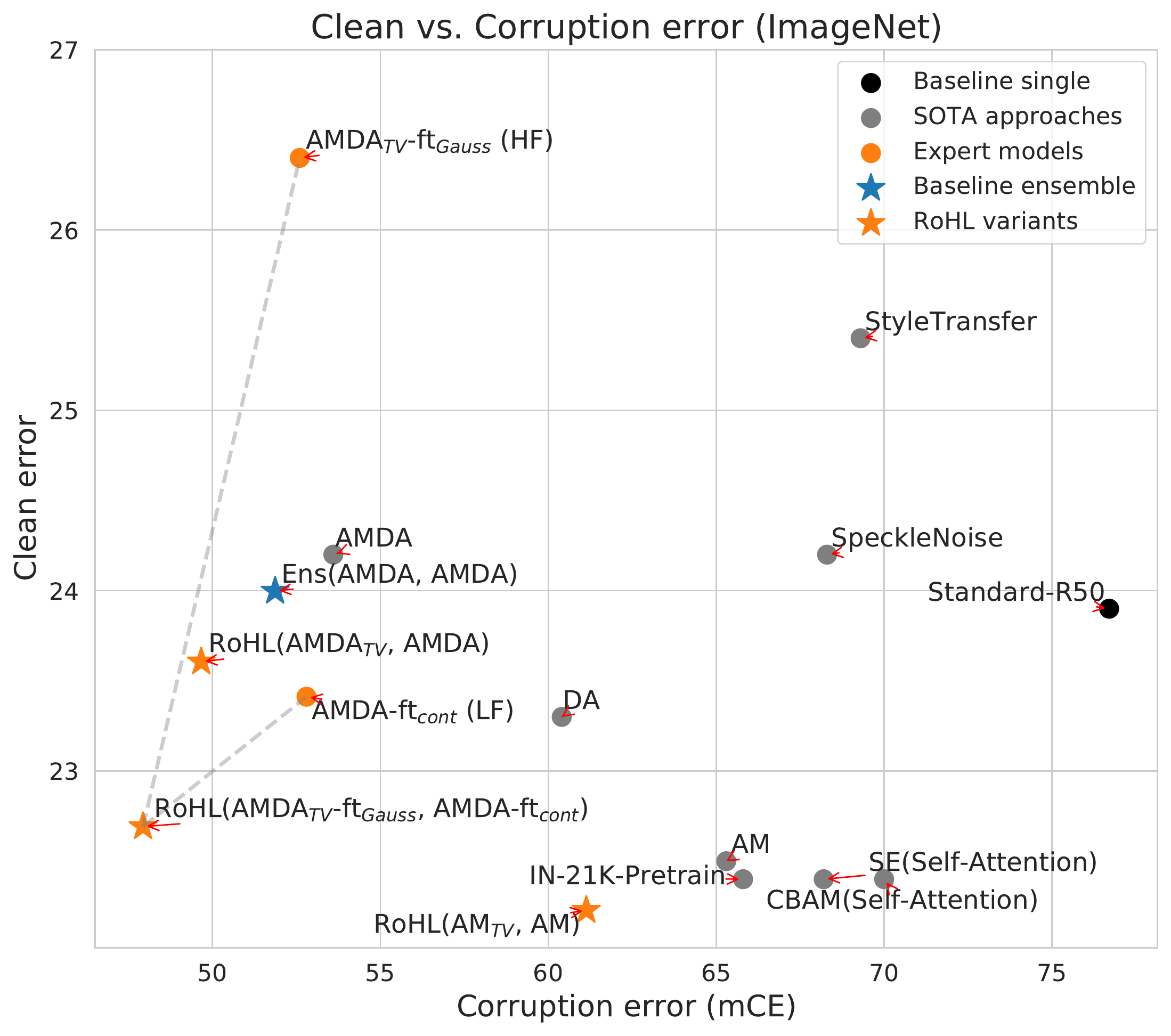}
\caption{Clean vs. corruption error on full ImageNet.
%Each circle represents the clean  (y-axis)  and  corruption  error  (x-axis) of a model. %Dots: Single models. Stars: Two member ensemble. Gray: Existing models. Blue: Ensemble baseline of two AMDA models, Orange: Our RoHF models.
The Pareto-front shows that our approach scales well and improves the previous state of the art on ImageNet-C.}
\label{fig:imagenet_pareto}
\end{center}
\end{figure}

\subsection{Results of RoHL variants}
In \tabref{tab:imagenet-sota-full} we compare performance of RoHL variants with other approaches and an ensemble of two AMDA models. We also show errors on each corruption type. The trade-off between Clean vs Corruption error is shown in \figref{fig:imagenet_pareto}. We observe that RoHL(\amdatvh, \amdal) outperforms the baseline Ensemble(AMDA, AMDA) on all high-frequency corruption types except Motion and Zoom blur. On low-frequency corruption types our approach performs the same or better. Also, we highlight that RoHL(\amdatv, \amda) also improves that state-of-the-art without any additional data augmentation.

\begin{table*}[h]
\centering
\caption{Detailed results on ImageNet and ImageNet-C. We compare RoHL to other state-of-the-art approaches using a ResNet50 architecture. We also compare to an ensemble of two AMDA models with already improves the state-of-the-art. RoHL shows the best trade-off between clean error and mCE. Individual errors for different corruption types are also shown. Error for each corruption type is normalized by AlexNet's error~\cite{hendrycks2019robustness} on that particular corruption. Therefore, values greater than 100 indicate worse performance compared to AlexNet.}
\label{tab:imagenet-sota-full}
\resizebox{\textwidth}{!}{%
\begin{tabular}{llll|ccc|cccc|cccc|cccc}
\toprule
  & & & & \multicolumn{3}{c}{Noise} & \multicolumn{4}{c}{Blurs} & \multicolumn{4}{c}{Weather} & \multicolumn{4}{c}{Digital}\\
  &                    Model & Clean err. &   mCE & Gauss. & Shot & Impulse & Defocus & Glass & Motion & Zoom & Snow & Frost &  Fog & Bright. & Cont. & Elastic & Pix. & JPEG \\
\midrule
                                      & Standard-R50 &       23.9 & 76.7 &            80 &        82 &           83 &          75 &        89 &         78 &       80 &  78 &   75 &  66 &        57 &      71 &               85 &      77 &              77 \\
                                      \hline 
\parbox[c]{1mm}{\multirow{10}{*}{\rotatebox[origin=c]{90}{\emph{SOTA approaches}}}} &                                    IN-21K-Pretraining &       22.4 & 65.8 &            61 &        64 &           63 &          69 &        84 &         68 &       74 &  69 &   71 &  61 &        53 &      53 &               81 &      54 &              63 \\
                           &     SE (Self-Attention) &       22.4 & 68.2 &            63 &        66 &           66 &          71 &        82 &         67 &       74 &  74 &   72 &  64 &        55 &      71 &               73 &      60 &              67 \\
                         &      CBAM (Self-Attention) &       22.4 & 70.0 &            67 &        68 &           68 &          74 &        83 &         71 &       76 &  73 &   72 &  65 &        54 &      70 &               79 &      62 &              67 \\
                          &      AdversarialTraining &       46.2 & 94.0 &            91 &        92 &           95 &          97 &        86 &         92 &       88 &  93 &   99 & 118 &       104 &     111 &               90 &      72 &              81 \\
                        &               SpeckleNoise &       24.2 & 68.3 &            51 &        47 &           55 &          70 &        83 &         77 &       80 &  76 &   71 &  66 &        57 &      70 &               82 &      72 &              69 \\
                         &             StyleTransfer &       25.4 & 69.3 &            66 &        67 &           68 &          70 &        82 &         69 &       80 &  68 &   71 &  65 &        58 &      66 &               78 &      62 &              70 \\
                          &                       AugMix &       22.5 & 65.3 &            67 &        66 &           68 &          64 &        79 &         59 &       64 &  69 &   68 &  65 &        54 &      57 &               74 &      60 &              66 \\
                        &                         DeepAugment &       23.3 & 60.4 &            49 &        50 &           47 &          59 &        73 &         65 &       76 &  64 &   60 &  58 &        51 &      61 &               76 &      48 &              67 \\
                         &                      AugMix+DeepAugment (AMDA) &       24.2 & 53.6 &            46 &        45 &           44 &          50 &        64 &         50 &       61 &  58 &   57 &  54 &        52 &      48 &               71 &      43 &              61 \\
                               \hline 
       \parbox[c]{1mm}{\multirow{4}{*}{\rotatebox[origin=c]{90}{\emph{Ours}}}}                   &          Ens(AMDA, AMDA) &       24.0 & 51.9 &            43 &        42 &           42 &          48 &        63 &         49 &       61 &  57 &   55 &  53 &        50 &      46 &               68 &      42 &              59 \\
       \cline{2-19}
                  &  RoHL(\amtv, \am) &      \bf{22.2} & 61.1 &            61 &        61 &           61 &          60 &        73 &         56 &       61 &  66 &   64 &  60 &        52 &      55 &               69 &      55 &              63 \\
                  &  RoHL(\amdatv, \amda) &       23.6 & 49.7 &            41 &        40 &           39 &          46 &        57 &         \bf{47} &       \bf{58} &  \bf{57} &   53 &  \bf{53} &        49 &      46 &               \bf{64} &      38 &              57 \\
  &  RoHL(\amdatvh,\amdal) &       22.7 & \bf{47.9} &            \bf{36} &        \bf{35} &           \bf{34} &          \bf{45} &  \bf{55} &         56 &       66 &  \bf{57} &   \bf{50} &  \bf{53} &        \bf{47} &     \bf{35} &              \bf{64} &      \bf{36} &            \bf{50} \\

\bottomrule
\end{tabular}

}
\end{table*}

%\section{Detailed analysis of Frequency biases}
\section{Results on BDD100k}
\begin{table}[h]
\centering
\caption{Object classification performance on BDD0100k-cls with ResNet50. We show errors on weather corruptions present in the BDD100k-cls test set. Corrupted samples are mostly benign and hence do not significantly degrade performance. }
\label{tab:bdd_cls}
\resizebox{0.9\columnwidth}{!}{
\begin{tabular}{l cc cc}
\toprule
Model  &    Clear  &  &      Rain & Snow \\
\cline{4-5}
       &    error & mCE  &  \multicolumn{2}{c}{errors}    \\
\hline
Standard data augmentation	&  \cellcolor{gray!15}5.8  & \cellcolor{gray!15}7.4     & 8.1 & 6.8 \\
AMDA                        &  \cellcolor{gray!15}5.3  & \cellcolor{gray!15}6.5     & 7.1 &  5.8\\
Ensemble(AMDA, AMDA)        &  \cellcolor{gray!15}5.1  & \cellcolor{gray!15}6.4     & 7.0 &  5.8\\
\rohl(\amdatvh,\amdal)      &  \cellcolor{gray!15}\bf{5.0} & \cellcolor{gray!15}\bf{6.2} & \bf{6.7} & \bf{5.6}\\
\bottomrule
\end{tabular}
}
\end{table}
\textbf{Object classification.} BDD100k is an object detection dataset containing multiple object instances per image and hence cannot be directly used in the classification setting. We extracted object images for each class using 2D bounding box annotations to first transform these datasets to the standard classification setting. The transformed variants are denoted as BDD100k-cls.  We finetuned our ResNet50 models (pre-trained on ImageNet) on the  "clear" split of BDD100k-cls. For RoHL, we finetune with the HF and LF biases. We evaluated on corrupted test sets of BDD100k-cls. We observed that weather distortions present in BDD100k are rather benign (see~\figref{fig:dataset_examples}). From \tabref{tab:bdd_cls} we observe that the corrupted test sets do not significantly impact performance of models trained even with standard data augmentation. %There is a $\sim\!\!2\%$ gap between \iid and \ood test sets.

\begin{table}[t]
\centering
\caption{Object detection performance with different ResNet50 backbones used in FasterRCNN on BDD100k. We report AP scores on the "Clear" and corrupted test splits of BDD100k. Higher AP scores are better. mAPc denotes the mean AP over corruption types.}
\label{tab:det-dawn}
\resizebox{0.9\columnwidth}{!}{
\begin{tabular}{lcccccc}
\toprule
      Pretrained backbone         &  Clear &   & Rain   &  Snow               \\
      \cline{4-5}
      &    AP & mAPc  &  \multicolumn{2}{c}{AP}    \\
\midrule
      Standard data augmentation &  \cellcolor{gray!15}~27.8       &  \cellcolor{gray!15}~25.6 &  27.6   &   23.6        \\
 %StyleTransfer                   &   27.7 &   28.0   &   24.2        &   26.1  \\
          AMDA                   &    \cellcolor{gray!15}~27.7       &  \cellcolor{gray!15}~25.7 &  27.4   &   23.9        \\
Ensemble(AMDA, AMDA)             &    \cellcolor{gray!15}~28.6       &  \cellcolor{gray!15}~26.6 & 28.5   &   24.7        \\
  \rohl(\amdatvh,\amdal)         & \cellcolor{gray!15}   \bf{28.7}  & \cellcolor{gray!15}~\bf{26.8} &   \bf{28.6}   &   \bf{25.0}    \\
\bottomrule
\end{tabular}
}
\end{table}

\textbf{Object detection.} Results for object detection are shown in \tabref{tab:det-dawn}.
We can observe that performance gap between \iid and \ood is marginal.

%\clearpage
\section{Results on other distribution shifts}
\subsection{ImageNet-R}
To measure performance on non-corruption based distribution shift we evaluate RoHL on ImageNet-R. We compare to other state-of-the-art approaches and a two-member ensemble of AMDA models.
We note that ImageNet-R contains a subset of 200 classes from ImageNet. Therefore to evaluate models trained on ImageNet we follow Hendryks~\etal~\cite{hendrycks2020many} and mask out predictions for irrelevant class indices. We do not train or finetune new models. The results are shown in \tabref{tab:imagenet-r}. \rohl improves on \iid and \ood test sets but the gains are diminished.

\begin{table}[h]
\centering
\caption{Results on ImageNet-200  and ImageNet-R. ImageNet-200 (IN-200) uses the same 200 classes as ImageNet-R (IN-R). Here IN-200 and IN-R are the \iid and \ood test sets respectively. \rohl improves both \iid and \ood performance compared to the state-of-the-art AMDA.}
\label{tab:imagenet-r}
\resizebox{0.8\columnwidth}{!}{%

\begin{tabular}{llcc}
\toprule
               &  Model & IN-200 & IN-R \\
               &           & error & error\\
\midrule
             & Standard ResNet50~\cite{he2016deep} &    7.9 &          63.9 \\
        \hline
\parbox[c]{1mm}{\multirow{8}{*}{\rotatebox[origin=c]{90}{\emph{SOTA approaches}}}} &
      IN-21K-Pretrain~\cite{hendrycks2020many} &      \bf{7.0} &          62.8 \\
&  CBAM(Self-Attention)~\cite{hendrycks2020many} &    \bf{7.0} &          63.2 \\
&   AdversarialTraining~\cite{wong2020fast} &   25.1 &          68.6 \\
 &         SpeckleNoise~\cite{rusak2020simple}  &    8.1 &          62.1 \\
 &        StyleTransfer~\cite{geirhos2018generalisation}  &    8.9 &          58.5 \\
 &               AM~\cite{augmix} &    7.1 &          58.9 \\
 &          DA~\cite{hendrycks2020many} &    7.5 &          57.8 \\
&        AMDA~\cite{hendrycks2020many}               &    8.0 &          53.2 \\
       \hline
\parbox[c]{1mm}{\multirow{3}{*}{\rotatebox[origin=c]{90}{\emph{Ours}}}}  &           Baseline Ensemble (AMDA, AMDA) &    8.0 &          52.3 \\
\cline{2-4}
% &  \rohl(\amdatv, \amda) &    7.8 &          \bf{51.4} \\
 &  \rohl(\amdatvh,\amdal) &    7.5 &          51.6 \\
\bottomrule
\end{tabular}
}
\end{table}
\subsection{ObjectNet}
We evaluate our ResNet50 models trained on ImageNet on ObjectNet's  test images. We excluded non-overlapping classes between ImageNet and ObjectNet. Results are shown in \tabref{tab:objectnet}. Considering the high baseline errors, the improvements are marginal.

\begin{table}[h]
\centering
\caption{Errors on ObjectNet with ResNet50. Lower is better.}
\label{tab:objectnet}
\resizebox{0.8\columnwidth}{!}{%

\begin{tabular}{lc}
\toprule
                Model & ObjectNet (error) \\
\midrule
Standard ResNet50      &   72.3  \\
AMDA                   &  72.4    \\
Ensemble(AMDA, AMDA)   &  72.3 \\
\rohl(\amdatvh,\amdal) &  \bf{70.8}\\
\bottomrule
\end{tabular}
}
\end{table}

\section{Dataset details}

\subsection{Visual examples of real image corruptions}
\begin{figure*}[h]
\begin{center}
\begin{subfigure}{0.48\linewidth}
        \includegraphics[width=\linewidth]{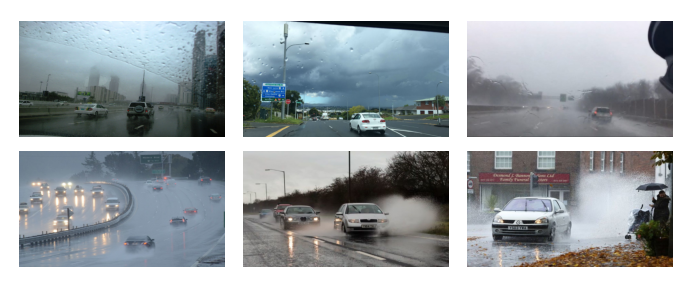}
        \caption{DAWN: Rain}
\end{subfigure}
\begin{subfigure}{0.48\linewidth}
        \includegraphics[width=\linewidth]{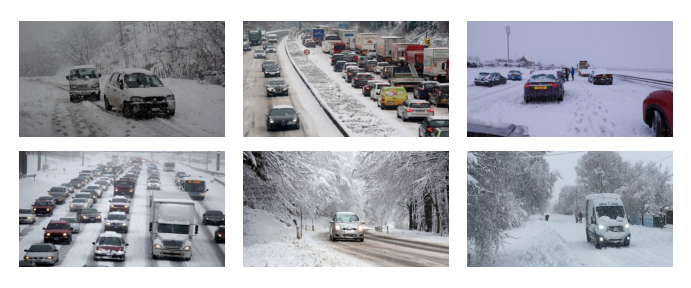}
        \caption{DAWN: Snow }
\end{subfigure}
\begin{subfigure}{0.48\linewidth}
        \includegraphics[width=\linewidth]{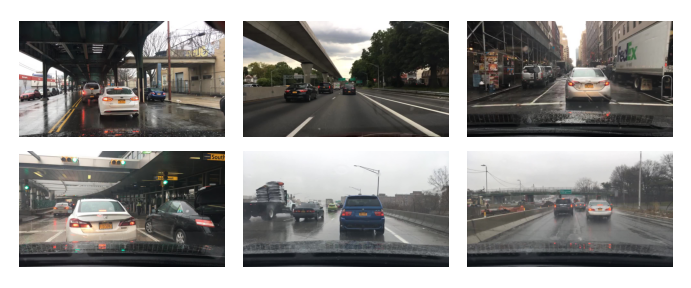}
        \caption{BDD100K: Rain}
\end{subfigure}
\begin{subfigure}{0.48\linewidth}
        \includegraphics[width=\linewidth]{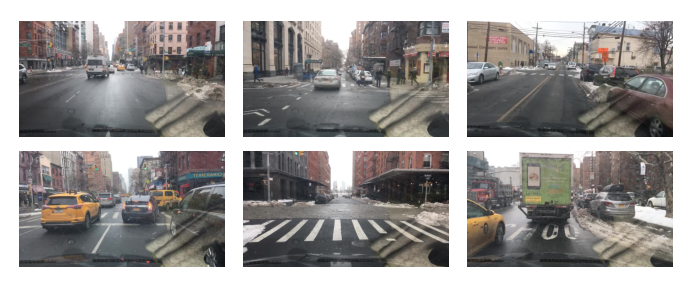}
        \caption{BDD100K: Snow}
\end{subfigure}
\caption{Example images of real image corruptions in BDD100k and DAWN. Images are randomly selected. DAWN contains more severe image corruptions and has a larger negative impact on \ood performance. }
\label{fig:dataset_examples}
\end{center}
\end{figure*}

\figref{fig:dataset_examples} shows example images of real image corruptions from BDD100k and DAWN.
We can observe that corrupted images on BDD100k are mostly benign. DAWN on the other hand contains more severe samples.

\subsection{ImageNet-100}
The class ids for the ImageNet-100 dataset used in our ablation studies are listed below:
\noindent
n01443537, n01484850, n01494475, n01498041, n01514859, n01518878, n01531178, n01534433, n01614925, n01616318, n01630670, n01632777, n01644373, n01677366, n01694178, n01748264, n01770393, n01774750, n01784675, n01806143, n01820546, n01833805, n01843383, n01847000, n01855672, n01860187, n01882714, n01910747, n01944390, n01983481, n01986214, n02007558, n02009912, n02051845, n02056570, n02066245, n02071294, n02077923, n02085620, n02086240, n02088094, n02088238, n02088364, n02088466, n02091032, n02091134, n02092339, n02094433, n02096585, n02097298, n02098286, n02099601, n02099712, n02102318, n02106030, n02106166, n02106550, n02106662, n02108089, n02108915, n02109525, n02110185, n02110341, n02110958, n02112018, n02112137, n02113023, n02113624, n02113799, n02114367, n02117135, n02119022, n02123045, n02128385, n02128757, n02129165, n02129604, n02130308, n02134084, n02138441, n02165456, n02190166, n02206856, n02219486, n02226429, n02233338, n02236044, n02268443, n02279972, n02317335, n02325366, n02346627, n02356798, n02363005, n02364673, n02391049, n02395406, n02398521, n02410509, n02423022

\end{document}